\documentclass[10pt,twocolumn,letterpaper]{article}

\usepackage[pagenumbers]{cvpr} 

\usepackage[dvipsnames]{xcolor}

\definecolor{cvprblue}{rgb}{0.21,0.49,0.74}
\usepackage[pagebackref,breaklinks,colorlinks,citecolor=cvprblue]{hyperref}

\usepackage[accsupp]{axessibility}
\usepackage{url}
\usepackage{comment}
\usepackage{amsmath,amssymb} 
\usepackage{multirow}
\usepackage{booktabs}
\usepackage{subcaption}
\usepackage{bm}
\usepackage{array}
\usepackage{xcolor}
\usepackage{graphicx}
\usepackage{enumitem}
\usepackage{makecell}
\usepackage{setspace}
\usepackage[symbol,hang,flushmargin]{footmisc}

\def\paperID{10127} 
\def\confName{CVPR}
\def\confYear{2024}

\title{\vspace{-0.5cm}UniHuman: A Unified Model For Editing Human Images in the Wild\vspace{-0.5cm}}

\author{
Nannan Li$^{1}$\footnotemark[1]{} \quad  
Qing Liu$^{2}$ \quad  
Krishna Kumar Singh$^{2}$ \quad
Yilin Wang$^{2}$ \\
Jianming Zhang$^{2}$ \quad
Bryan A. Plummer$^{1}$ \quad
Zhe Lin$^{2}$ \\
$^{1}$Boston University \quad
$^{2}$Adobe \\
{\tt\small nnli@bu.edu \quad \{qingl,krishsin,yilwang,jianmzha\}@adobe.com }\\
{\tt\small bplum@bu.edu \quad zlin@adobe.com}
\vspace{-0.3cm}
}

\newcommand{\Ours}{UniHuman}
\newcommand{\rulesep}{\unskip\ \vrule\ }
\renewcommand{\paragraph}[1]{ \noindent \textbf{#1.}}
\AtBeginEnvironment{equation}{
\setlength{\abovedisplayskip}{3.5pt}
\setlength{\belowdisplayskip}{3.5pt}
}
\AtBeginEnvironment{equation*}{
\setlength{\abovedisplayskip}{3.5pt}
\setlength{\belowdisplayskip}{3.5pt}
}
\AtBeginEnvironment{gather}{
\setlength{\abovedisplayskip}{3pt}
\setlength{\belowdisplayskip}{3pt}
}

\begin{document}

\twocolumn[{%
\renewcommand\twocolumn[1][]{#1}%
\maketitle
\begin{center}
\vspace{-0.55cm}
\includegraphics[width=0.9\textwidth]{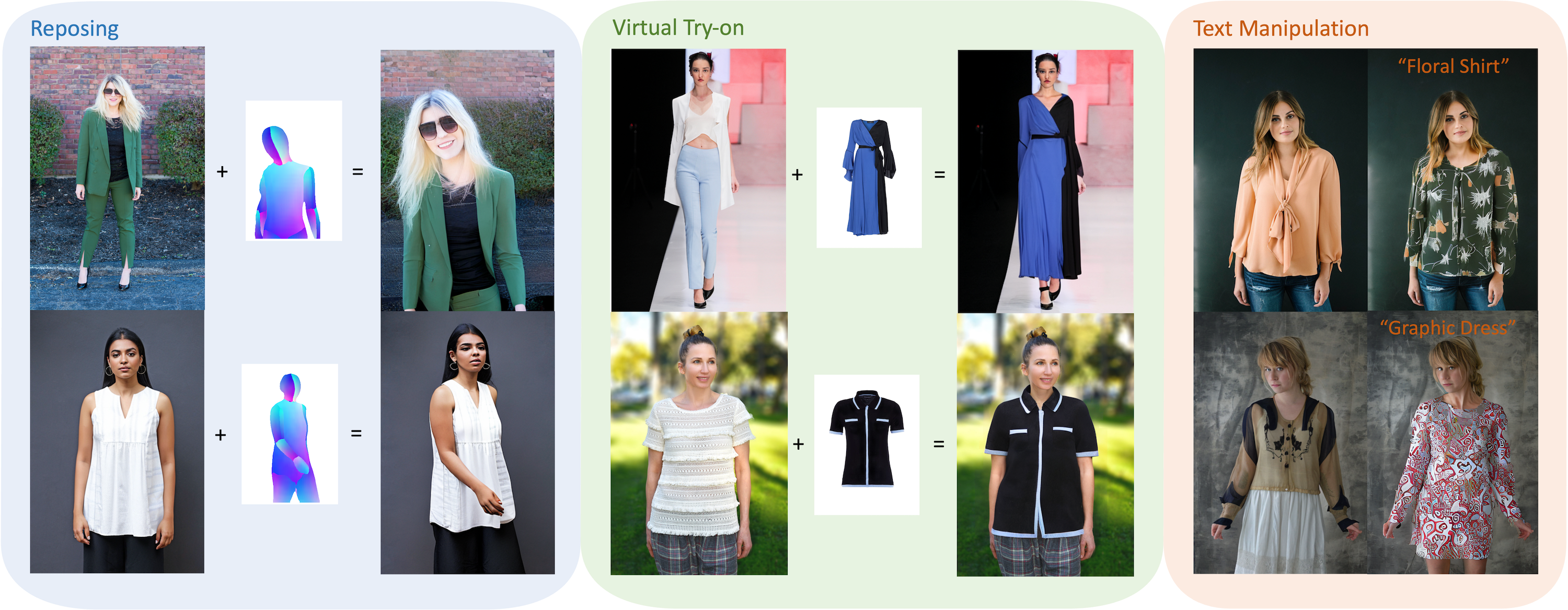}
\vspace{-0.3cm}
    \captionof{figure}{The results of UniHuman on diverse real-world images. UniHuman learns informative representations by leveraging multiple data sources and connections between related tasks, achieving high-quality results across various human image editing objectives.}  
    \label{fig:1}
\end{center}
}]

\footnotetext[1]{Work during Nannan Li's 2023 summer internship at Adobe Research.}


\begin{abstract}
\vspace{-6pt}
Human image editing includes tasks like changing a person's pose, their clothing, or editing the image according to a text prompt. However, prior work often tackles these tasks separately, overlooking the benefit of mutual reinforcement from learning them jointly. In this paper, we propose \Ours{}, a unified model that addresses multiple facets of human image editing in real-world settings. To enhance the model's generation quality and generalization capacity, we leverage guidance from human visual encoders and introduce a lightweight pose-warping module that can exploit different pose representations, accommodating unseen textures and patterns. Furthermore, to bridge the disparity between existing human editing benchmarks with real-world data, we curated 400K high-quality human image-text pairs for training and collected 2K human images for out-of-domain testing, both encompassing diverse clothing styles, backgrounds, and age groups. Experiments on both in-domain and out-of-domain test sets demonstrate that \Ours{} outperforms task-specific models by a significant margin. In user studies, \Ours{} is preferred by the users in an average of 77\% of cases. Our project is available at \href{https://github.com/NannanLi999/UniHuman}{this link}.
%


\end{abstract}

\section{Introduction}
In the realm of computer graphics and computer vision, the synthesis and manipulation of human images have evolved into a captivating and transformative field. This field holds invaluable applications covering a range of domains: reposing strives to generate a new pose of a person given a target pose \citep{pidm,zhang2021pise,zhang2022exploring,zhou2022cross}, virtual try-on aims to seamlessly fit a new garment onto a person \citep{ladiviton,zhu2023tryondiffusion,li2023virtual}, and text-to-image editing manipulate a person's clothing styles based on text prompts \citep{chang2023muse,gao2023editanything,goel2023pair,yang2023paint}. However, most approaches address these tasks in isolation, neglecting the benefits of learning them jointly to mutually reinforce one another via the utilization of auxiliary information provided by related tasks \citep{crawshaw2020multi,multi2,multi3}. In addition, few studies have explored effective ways to adapt to unseen human-in-the-wild cases. 


In response to these challenges, our goal is to unify multiple human image editing tasks in a single model, boosting performance in all settings. Thus, we propose \Ours{}, a unified model that exploits the synergies among reposing, virtual try-on, and text manipulation for in-the-wild human image editing. \cref{fig:1} shows \Ours{}'s high-quality generations can both preserve the clothing identity and recover texture details. While prior work \citep{cheong2023upgpt} made initial attempts to bridge the visual and text domains, connections within visual tasks remain largely under-explored. In contrast, our model takes a step further by exploiting the relationship between reposing and virtual try-on.  Specifically, reposing requires modifying the pose of all body parts and clothing items, while virtual try-on only adapts the pose of the target garment. Importantly, both tasks need to keep the visible texture consistent, either from the source image or the target garment, consistent after the pose or garment change. This suggests learning the two tasks could benefit each other when consolidated in a unified model. Moreover, when adding text prompts \citep{ladiviton}, the semantic information learned from large language models \citep{stablediff} could help our model synthesize more realistic textures based on the visible texture. 

Recognizing that all three tasks should maintain the consistency of visible content, we introduce a pose-warping module. Instead of being a task-specific module trained from a single format of pose representation \citep{grigorev2019coordinate,albahar2021pose,li2023virtual,xie2023gp,hrviton,vitonhd,yang2020towards}, our pose-warping module can explicitly leverage both dense and sparse pose correspondences to obtain visible pixels on all three tasks, equipping it with the capacity to handle previously unseen textures and patterns. 
In parallel, to maximize the utilization of human-specific visual information, we leverage human visual encoders at both the part and global levels to infuse texture, style, and pose guidance in the generation process.
The visual representations are then taken as reference in the unified model to reconstruct the person after the desired pose/garment change.
Our experiments show that the introduced pose-warping module can enhance the model's generalization capacity as well as generation quality at inference. 

In order to adapt our model to real-world scenarios, it is essential to have ample data for learning informative representations across diverse domains. However, existing datasets are constrained by the scale and diversity \citep{liu2016deepfashion,jiang2022text2human,dresscode,vitonhd}, particularly in terms of pose, background, and age groups. Relying solely on such data for training may introduce biases in image generation and hinder the model's ability to generalize to real-world samples. Additionally, existing in-the-wild human image/video datasets \citep{tiktok,liu2016deepfashion} are marred by motion blur or by a lack of effective way to handle occlusions. Instead, we developed an automated data-curation pipeline to collect 400K high-quality human image-text pairs from LAION-400M \citep{schuhmann2021laion400m}, assembling images with at least 512x512 resolution as well as little to no human-object occlusion. This cost-effective new dataset contains single-human images with an extensive range of poses, clothing styles, backgrounds, \emph{etc}. By jointly training on this dataset with existing datasets, we effectively enhance our model's generalization capacity on human-in-the-wild cases. Our contributions are:


\begin{itemize}
    \item We propose \Ours{}, a unified model that can achieve multiple in-the-wild human image editing tasks through human visual encoders. The incorporation of the pose-warping module further enhances the model’s generalization capability at inference.
    \item We curated 400K high-quality human image-text pairs for training and collected 2K human image pairs for out-of-domain testing. Our data expands existing data and includes more diverse poses,  backgrounds, and age groups. 
    \item Extensive experiments on both in-domain and out-of-domain test sets show that \Ours{} outperforms task-specific models quantitatively and qualitatively, and is preferred by the users in 77\% cases on average. 
\end{itemize}

\begin{figure*}
    \centering
    \begin{subfigure}[b]{0.68\textwidth}
        \centering
        \includegraphics[width=\textwidth]{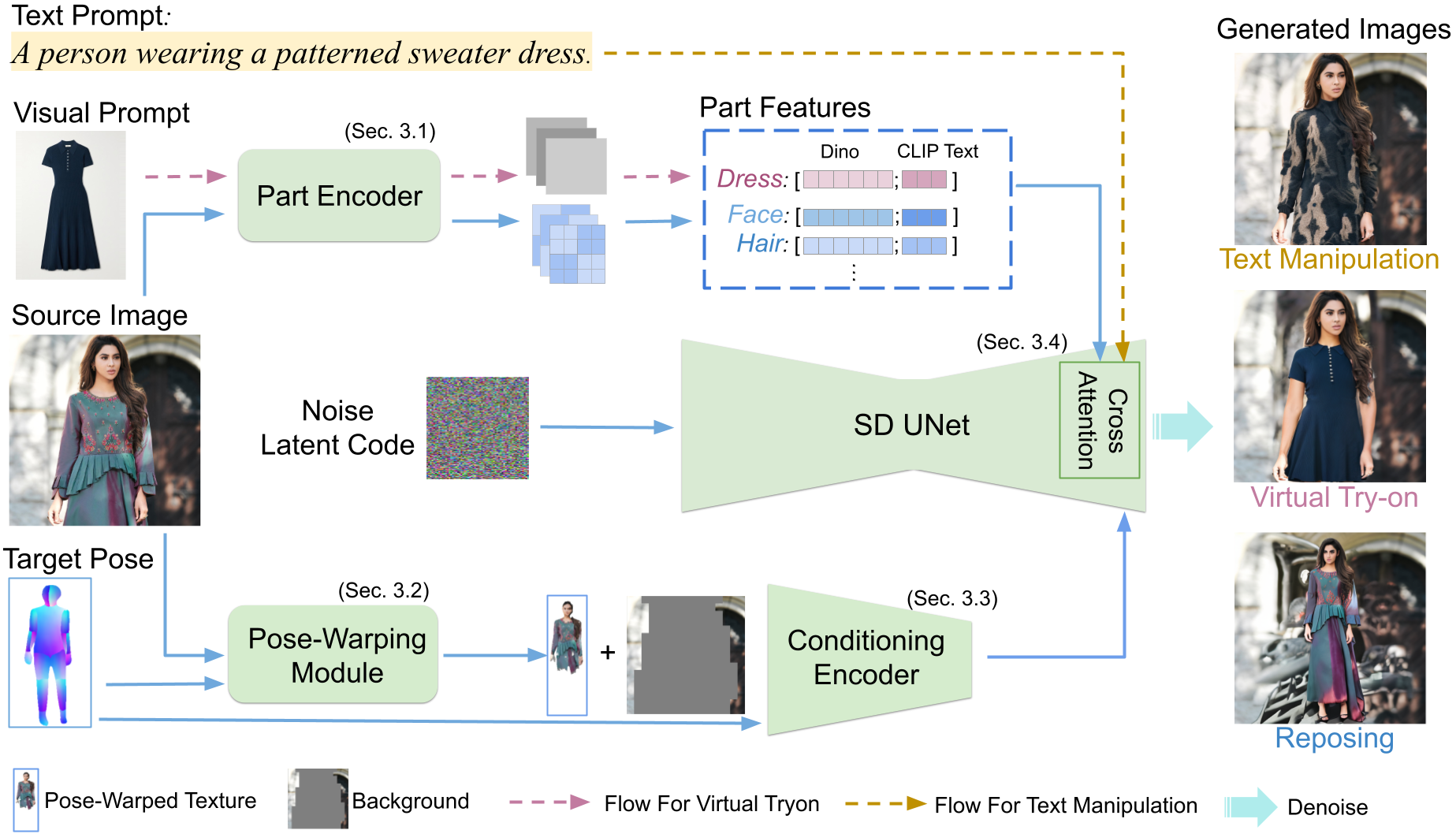}        
        \caption{Inference pipeline (\cref{sec:framework}).}
        \label{fig:main model}
    \end{subfigure}
    \hfill
    \begin{subfigure}[b]{0.29\textwidth}
        \centering
        \includegraphics[width=\textwidth]{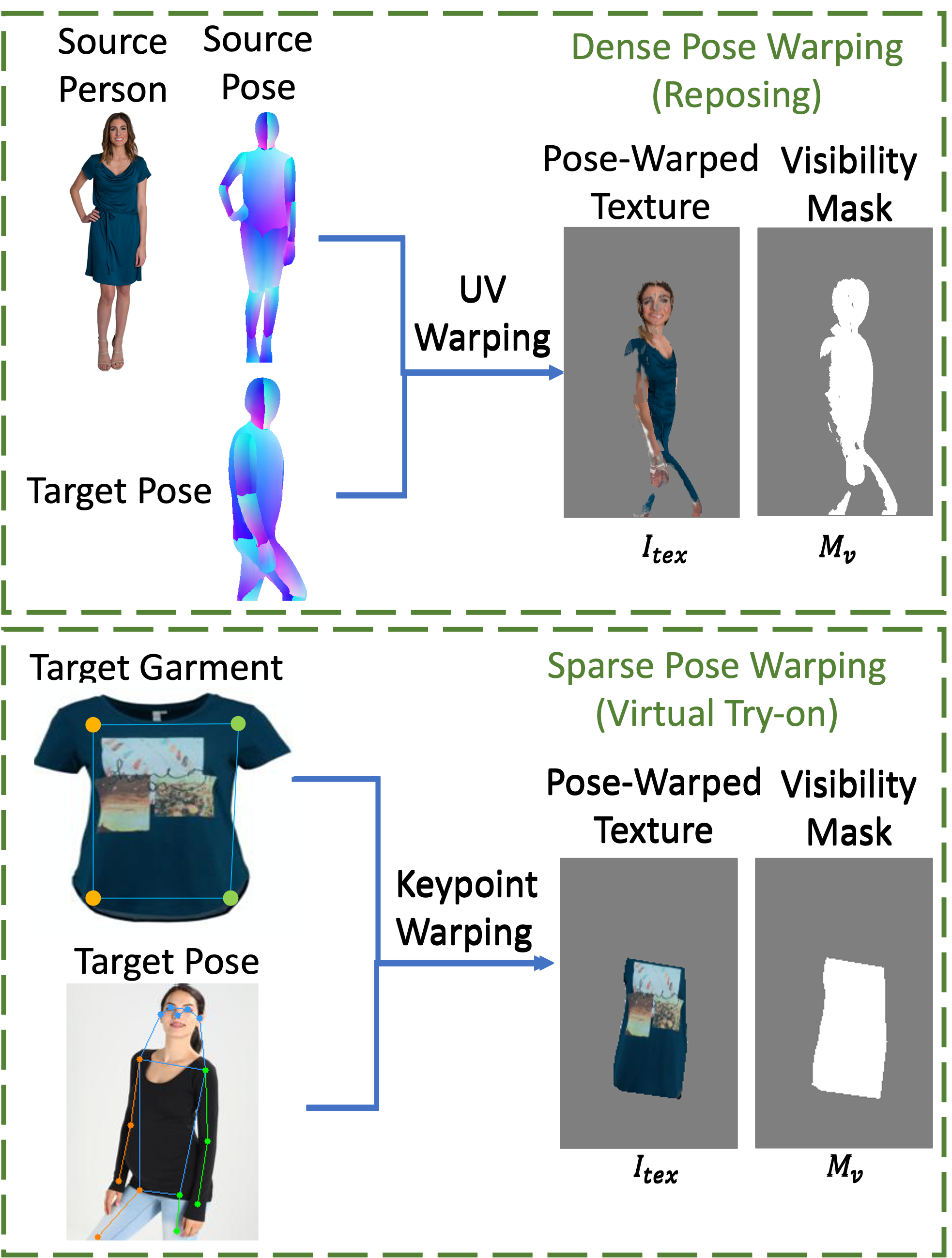}       
        \caption{Pose-Warping module. (\cref{sec:pw tex})}
        \label{fig:pw tex}
    \end{subfigure}
    \caption{An overview of our model. \textbf{(a)} Our inference pipeline. Starting from a noise latent code, our model edits the source person given the source image, the target pose, the visual prompt (optional), and the text prompt (optional). \textcolor{blue}{Blue arrow} is the reposing flow, which is also the base flow for all tasks. \textcolor{pink}{Pink dashed arrow} indicates the optional virtual try-on flow that takes a clothing image as its input. In \emph{try-on} task, the clothing image should replace the source image as the input to the pose-warping module. \textcolor{brown}{Brown dashed arrow} is the optional text manipulation flow, which accepts a text description as its prompt. \textbf{(b)} The introduced pose-warping module. It maps the original RGB pixels of the source texture to the target pose based on pose correspondences. Best view in color.}
    \label{fig:model}
    \vspace{-0.2cm}
\end{figure*}
\section{Related Work}
Human image editing empowers us to reshape, recolor, and infuse texture patterns into visual content. This paper tackles three distinct yet interrelated tasks: reposing, virtual try-on, and text manipulation. In contrast to methods that handle these tasks independently \citep{pidm,cheong2023upgpt,albahar2021pose,wang2023disco,dreampose,zhang2022exploring,Cui_2021_ICCV,wang2022self,li2023collecting,pasta,ladiviton,zhu2023tryondiffusion,hrviton,li2023virtual,xie2023gp,chang2023muse,goel2023pair,yang2023paint,gao2023editanything}, our unified model, \Ours{}, leverages synergies and mutual benefits from related tasks through joint learning. Additionally, we augment existing datasets with additional diverse and high-quality data to enable our model's adaptability to unseen human-in-the-wild cases, a less-explored setting in prior research.

Previous attempts at unifying the three tasks \citep{cheong2023upgpt} used frozen CLIP embeddings \citep{clip} to bridge visual and text domains. However, the connections between different \textit{visual} tasks in this setting still remained largely overlooked. In contrast, our model unifies all three human image editing tasks by introducing a pose-warping module. This module provides texture details derived from pose aligning, equipping it with the ability to handle unseen clothing patterns. 
While other methods \citep{grigorev2019coordinate,albahar2021pose,li2023virtual,xie2023gp,hrviton,vitonhd,yang2020towards} often 
rely on a single format of pose representation to train task-specific pose warping modules, \Ours{} flexibly utilizes both dense and sparse pose representations for training-free texture warping on all three tasks. This approach mitigates the overfitting to specific datasets, thus ensuring accurate visible pixel information resilient to domain shifts. More related literature discussions can be found in Supp.

\section{Human Image Editing in a Unified Model}
\label{sec:framework}

Given a source image $I_{src}$, a target pose $P_{tgt}$, an optional visual prompt $G_t$ and/or an optional text prompt $y$, our goal is to reconstruct the person of $I_{src}$ at the target pose $P_{tgt}$, and simultaneously transfer the texture from $G_t$ and/or create a new texture based on $y$. Note that we unify three human image editing tasks in one single model as follows: In the absence of a visual/text prompt, our focus narrows down to human reposing; when the visual prompt specifies a target garment, the task transforms into virtual try-on; conversely, when guided by a text description, the model edits the image following the principle of text manipulation. 

Our \Ours{} model is implemented based on Stable Diffusion (SD) \citep{stablediff}. The part encoder learns texture styles from segmented human parts, supplying them to SD cross-attention (\cref{sec:part enc}). Simultaneously, the pose-warping module generates detailed, target pose-aligned visible texture (\cref{sec:pw tex}). These outputs, along with the target pose and partial background, channel into the SD UNet via a conditioning encoder (\cref{sec:cond enc}). For virtual try-on (\cref{sec:flows}), the optional target garment is injected into the part encoder to be combined with other human parts. In cases of text manipulation, the SD UNet learns semantic information from an optional text prompt (\cref{sec:flows}). After $N$-timestep denoising and VAE decoding, the model produces a clean edited image. Objective functions used in training are detailed in \cref{sec:loss}.

\subsection{Part Encoder}
\label{sec:part enc}
To acquire texture information from the source person, we utilize a part encoder to obtain segmented human part features, which are then fed into the SD UNet decoder. Unlike the approach in \citet{cheong2023upgpt}, where human parts are segmented at the pixel level and encoded separately, we segment parts at the feature level, \ie, take parts from the feature map of the entire source person. We find this segmented feature map can preserve more contextual information than image segments, \eg, the length of the clothing and interactions between the upper \& lower clothing. An off-the-shelf human parsing model \citep{parser} is used to extract \textit{face, hair, headwear, upper clothing, coat, lower clothing, shoes, accessories}, and \textit{person} from the source person's DINOv2 \citep{dinov2} feature map. These visual features are then concatenated with the corresponding CLIP text embeddings. Let $d_{\omega}$ be the part encoder that includes DINOv2 and CLIP. The obtained part features $B=d_{\omega}(I_{src})$ will provide source texture and style information in the SD UNet in \cref{sec:flows}.


\subsection{Pose-Warping Module}
\label{sec:pw tex}
To ensure texture consistency after pose/garment change and improve our model's ability to generalize to unseen textures, we introduce a pose-warping module. This module produces the pose-warped texture $I_{tex}$ and its binary mask $M_v$, which will be sent to the conditioning encoder in \cref{sec:cond enc} and to the SD UNet cross-attention in \cref{sec:flows}.
While earlier approaches \citep{grigorev2019coordinate,albahar2021pose,li2023virtual,li2023virtual,xie2023gp,hrviton,yang2020towards} train task-specific pose warping modules, our model obtains the pose-warped texture through explicit correspondence mapping as shown in \cref{fig:pw tex}. This process only needs an off-the-shelf pose detector to provide sparse or dense pose prediction for texture warping. Consequently, our method is inherently more resilient to domain shifts across different tasks, achieving enhanced generalization capacity to handle unseen patterns and styles.


For tasks involving human pose change, the pose-warped texture $I_{tex\text{-}{rp}}$ pertains to pixels that remain visible after reposing, as shown in \cref{fig:pw tex} top panel. We use UV map correspondence to resample source RGB pixels such that they are aligned with the target pose. This alignment is critical in enabling direct reconstruction of intricate texture patterns. However, in cases where only the target garment requires repositioning (as in virtual try-on), since no 3D or contextual information is available from the target garment image, warping the texture through UV coordinates becomes unfeasible. In such scenarios, as illustrated in \cref{fig:pw tex} bottom panel, we pivot to the use of sparse keypoints to apply a perspective warping from the canonical view of the target garment to the human torso. This warping repositions the clothing texture to the desired pose, providing the pose-warped texture $I_{tex\text{-}vt}$ for virtual try-on. For text manipulation, the pose-warped texture $I_{tex\text{-}{tm}}$ exhibits adaptability, catering to user-specific requirements. For example, it can be set to zero to facilitate the generation of clothing textures from scratch based on the text input. Our experiments show that the introduced pose-warped texture strengthens the generalization capacity of our approach.

\subsection{Conditioning Encoder}
\label{sec:cond enc}

The conditioning encoder takes the target pose $P_{tgt}$, pose-warped texture $I_{tex}$ and partial background ${I_{bg}}$ as input, which provides essential posture guidance and visible texture reference for all tasks. The partial background image ${I_{bg}}$ is extracted by masking out the bounding boxes of the source and target pose region.
Following \citep{controlnet}, the encoded features in $g_{\phi}$ are concatenated with the intermediate features in SD UNet decoder as follows
\begin{equation}    
    \hat{h}^i = W^i_h [h^i;g_{\phi}^i([I_{tex};P_{tgt};{I_{bg}}])],
    \label{eq:cond_enc}
\end{equation}

\noindent
where $h^i$ is the $i^\text{th}$ intermediate feature map of the SD UNet decoder, $g_{\phi}^i$ is the $i^\text{th}$ intermediate layer of $g_{\phi}$, $W_h^i$ are learnable weights, and $[\cdot;\cdot]$ indicates concatenation. To obtain $\hat{h}^i$, the output feature maps of $g_{\phi}$ at varying resolutions are injected into every block of the SD UNet decoder. We define $E=g_{\phi}([I_{tex};\emptyset;\emptyset])$ as the encoded pose-warped texture by itself in the last layer of $g_{\phi}$, which will be sent to the SD UNet cross-attention described by \cref{eq:AttRP} in \cref{sec:flows} to further improve the texture quality. 

\subsection{Image Editing Pipelines}
\label{sec:flows}

We exemplify our pipeline using reposing. Following the \textcolor{blue}{blue arrows} in \cref{fig:main model}, the denoising process is guided by a target pose, which will be enriched by textures from the source person. The texture information has two sources: the part features $B$ in \cref{sec:part enc}, and the pose-warped texture $I_{tex}$ in \cref{sec:pw tex}. The part features $B$ preserve style information, helping maintain the overall authenticity of the generated clothing, and $I_{tex}$ provides detailed and spatial aligned textures, ensuring high fidelity in the generated image. 

With $B$ and $I_{tex}$ serving as the texture sources, their information is transmitted by a cross-attention with the intermediate layers of SD UNet decoder:
\begin{gather}
\text{Attention}(Q,K,V)=\text{softmax}(QK^T) \cdot V, \label{eq:crossAtt} \\
    Q^i=W^i_Q h^i,\text{ }K^i=W^i_K [B;E],\text{ }V^i=W^i_V [B;E], \label{eq:AttRP}
\end{gather}
\noindent
where $h^i$ is the $i^\text{th}$ intermediate feature representation of SD UNet decoder. $W^i_Q,W^i_K$ and $W^i_V$ are learnabled weights. $E$ indicates the encoded pose-warped texture in the conditioning encoder. In the following, we use $E_{rp},E_{vt}$ and $E_{tm}$ to denote the encoded pose-warped texture for each task. 


Finally, with the SD denoising function $f_{\theta}$, we obtain the latent code for reposing $I_{rp}^{(t)}$ at time step $t$ by
\begin{equation*}
    I^{(t)}_{rp}=f_{\theta} \Bigl(g_{\phi}([I_{tex\text{-}rp};P_{tgt};{I_{bg}}]),B,E_{rp},I_{rp}^{(t+1)},y \Bigr),
\end{equation*}
\noindent
where $y$ is the optional text prompt that will also be mapped to the UNet decoder via the standard cross-attention block in SD \citep{stablediff}. This text cross-attention is applied after the part cross-attention in \cref{eq:crossAtt,eq:AttRP}. 


In virtual try-on (\textcolor{pink}{pink dashed arrows} in \cref{fig:main model}), the source garment $G_s$ is first removed and then replaced by the target garment $G_t$ in the part features. Let $I_{src} - G_s$ be the image without the source garment. The part features in virtual try-on thus becomes $B^{\prime}=[d_{\omega}(I_{src} - G_s);d_{\omega}(G_t)]$. $B^{\prime}$ is then utilized in denoising as 
\begin{equation*}
    {I}_{vt}^{(t)}=f_{\theta}\Bigl( g_{\phi}([I_{tex\text{-}vt};P_{src};{I_{bg}}]),B^{\prime},E_{vt},I_{vt}^{(t+1)},y \Bigr),
\end{equation*}
\noindent
where the source pose $P_{src}$ is used as guidance since virtual try-on doesn't change the original posture of the person.

 Our model can also be used to edit the garment according to a text prompt (\textcolor{brown}{brown dashed arrows} in \cref{fig:main model}). Similar to virtual try-on, the described source garment $G_s$ is removed from the source image $I_{src}$, for which we get $B^{\prime}$. The garment's missing information will be replenished by the text cross-attention in SD, resulting in the following denoising process, 
 \begin{equation*}
    {I}_{tm}^{(t)}=f_{\theta}\Bigl( g_{\phi}([I_{tex\text{-}tm};P_{src};{I_{bg}}]),B^{\prime},E_{tm},I_{tm}^{(t+1)},y \Bigr).
\end{equation*}
\subsection{Objective Functions}
\label{sec:loss}

To prevent the texture blending problem \citep{xiao2023fastcomposer}, we apply two loss functions to constrain the cross attention for different human parts and pose-warped texture.
For each human part $p_n$, let $A_{p_n}$ and $M_{p_n}$ be the attention map of $B_{p_n}$ and segmentation map (resized to the same size), respectively. Following \citep{xiao2023fastcomposer}, we minimize their distance by:
\begin{equation*}
    \mathcal{L}_{B}=\sum\nolimits_{n} \Bigl( \text{mean}(A_{p_n} \odot (1-{M}_{p_n}))-\text{mean}(A_{p_n} \odot M_{p_n}) \Bigr).
    \label{eq:loss_b}
\end{equation*}
Similarly, for the pose-warped texture, using the binary visibility map obtained from \cref{sec:pw tex}, we constrain the attention map of $E$ by:
\begin{equation*}
    \mathcal{L}_{E}=\text{mean}(A_{v} \odot (1-{M}_{v}))-\text{mean}(A_{v} \odot M_{v})
    \label{eq:loss_e}.
\end{equation*}

Driven by the two losses, our model is steered towards sampling from the pose-warped textures for visible pixels, and from part features for invisible regions. The net result is a harmonious interplay that ensures accurate reconstruction and optimized fidelity for the entire generated content.

For standard SD training loss, in practice, the UNet predicts a noise $\epsilon_{\theta}$ given the noisy version of the target image $I_{tgt}^{(t)}$ at each time step $t$. Let $\epsilon$ be the ground-truth noise and the L2 loss function is simplified as
\begin{equation*}
    \mathcal{L}_{SD}=\mathbb{E}_{I_{tgt},\epsilon \sim \mathcal{N}(0,1),t} \| \epsilon - \epsilon_{\theta}(I_{tgt}^{(t)},t,...) \|^2_2,
\end{equation*}
where $...$ omits other conditional inputs in our model, including $B$, $P_{tgt}$ (or $P_{src}$), $I_{bg}$, $I_{tex}$ and $y$. In summary, the overall objective function of our model is
\begin{equation*}
    \mathcal{L}=\mathcal{L}_{SD}+ \lambda_1 \mathcal{L}_{B} + \lambda_2 \mathcal{L}_{E},
\end{equation*}
where $\lambda_1$ and $\lambda_2$ are trade-off parameters.

\begin{figure}[!t]
    \centering
    \begin{subfigure}[c]{0.25\textwidth}
        \centering
        \includegraphics[width=\textwidth]{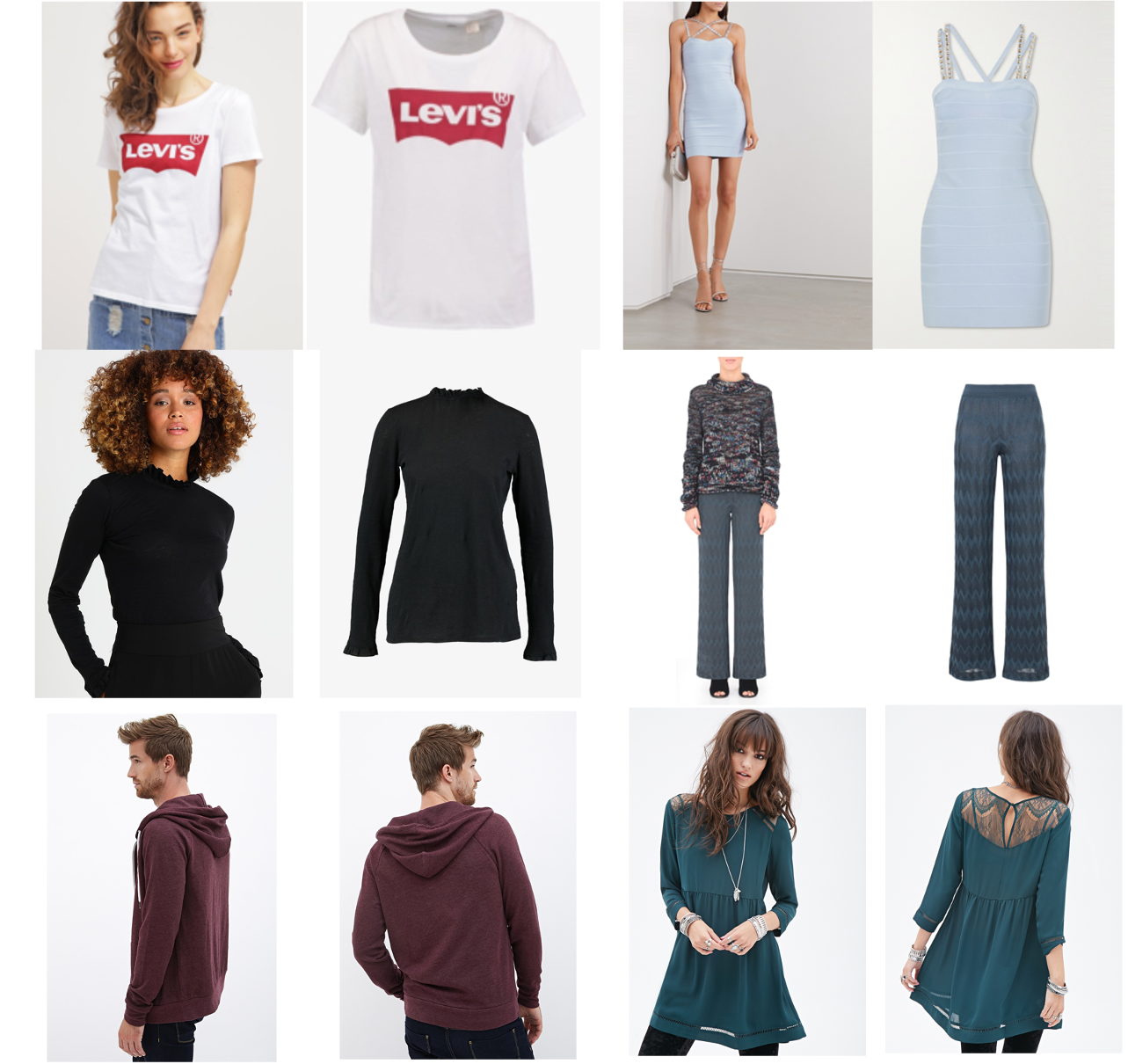}        
        \caption{Existing datasets \citep{liu2016deepfashion,jiang2022text2human,dresscode,vitonhd}.}
        \label{fig:curr_data}
    \end{subfigure}
    \hfill
    \begin{subfigure}[c]{0.21\textwidth}
        \centering
        \includegraphics[width=\textwidth]{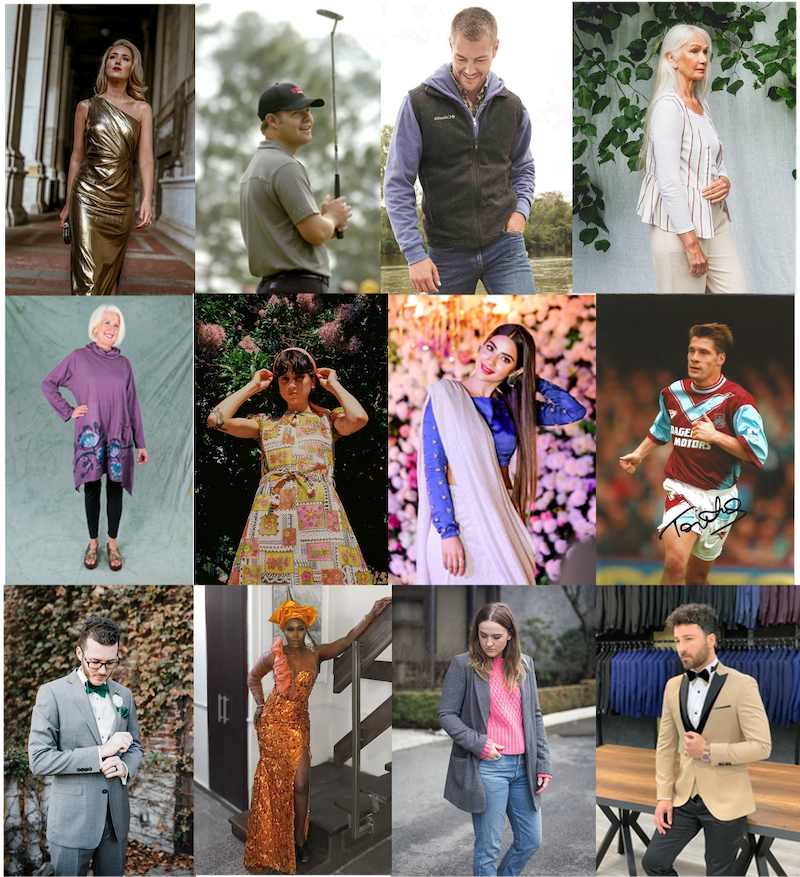}       
        \caption{LH-400K.}
        \label{fig:lh-400k}
    \end{subfigure}
    \vspace{-0.2cm}
    \caption{Representative examples from different datasets. Our LH-400K includes people of diverse ages and backgrounds.}
    \label{fig:compare_lhuman}
    \vspace{-0.1cm}
\end{figure}

\section{Collecting High-Quality Human Images}
\label{sec:data}

Exiting datasets for human image editing exhibit limitations in scale and diversity since most of these images were in-studio photography with fashion models (\eg,\citep{liu2016deepfashion,jiang2022text2human,dresscode,vitonhd}). This results in images with simple indoor backgrounds and younger age groups, as depicted in \cref{fig:curr_data}. To address this, we meticulously curate a larger-scale training dataset with augmented diversity (\cref{sec:train data}). Additionally, we collect new challenging test sets for out-of-domain evaluation of existing models (\cref{sec:test data}).

\begin{table}[!t]
    \centering
    \small
    \setlength{\tabcolsep}{2.5pt}
    \begin{tabular}{@{}lcccc@{}}
    \toprule
    &Dataset & \makecell{\#Train \\ Images} & \makecell{\#Test\\ Pairs}  & Background \cr
    \midrule  
     \multirow{4}{*}{In-Domain}
     &DeepFashion \citep{jiang2022text2human} &44,096 &8,512 &Gray\&White \cr  
     &DressCode \citep{dresscode}  &96,784 &5,400  &Walls \cr 
     &VITON-HD \citep{vitonhd}  &23,294  &2,032  &Gray\&White \cr 
     &LH-400K (Ours) &409,270 &- &Diverse \cr
     \midrule
     \multirow{2}{*}{Out-Of-Domain}
     &WPose (Ours) &- &2,304 &Diverse \cr 
     &WVTON (Ours) &- &440 &Diverse \cr 
    \bottomrule
    \end{tabular}
    \vspace{-0.2cm}
    \caption{ Statistics for the training and evaluation sets used in our experiments. LH-400K provides large-scale diverse human data for training (\cref{sec:train data}), while WPose and WVTON are collected to evaluate the out-of-domain generalization capacity (\cref{sec:test data}).}
    \label{tab:data}
    \vspace{-0.2cm}
\end{table} 

\subsection{Expanding the Training Data}
\label{sec:train data}
We introduce LH-400K, a large-scale dataset of high-quality single-human images selected from LAION-400M \citep{schuhmann2021laion400m} with diverse backgrounds, age groups, and body shapes. Notable distinctions between LH-400K and existing human image editing benchmarks are presented in \cref{fig:compare_lhuman} and \cref{tab:data}. To build LH-400K, we follow several criteria and build an automated pipeline to filter clean human images from the large noisy pool. By combining the existing modest-scale paired data with our newly introduced large-scale unpaired data during training, our model can better adapt to diverse real-world scenarios. See Supp. for details of how we incorporate LH-400K in our model training.

\begin{figure*}[!t]
    \centering
    \begin{subfigure}[c]{0.43\textwidth}
        \centering
        \includegraphics[width=\textwidth]{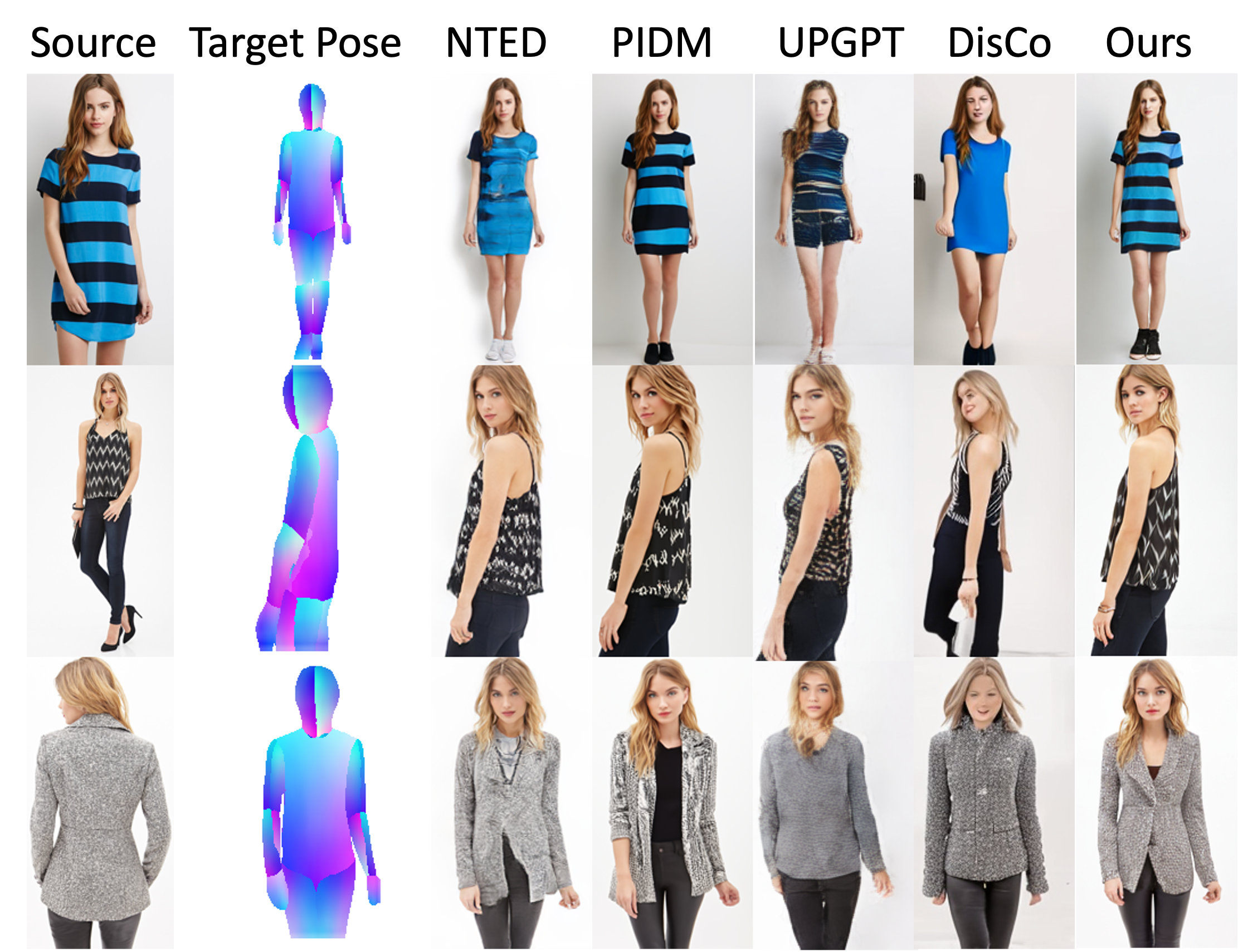}        
        \caption{In-domain examples on DeepFashion \citep{jiang2022text2human}.}
        \label{fig:iid_reposing}
    \end{subfigure}
    \rulesep
    \begin{subfigure}[c]{0.55\textwidth}
        \centering
        \includegraphics[width=\textwidth]{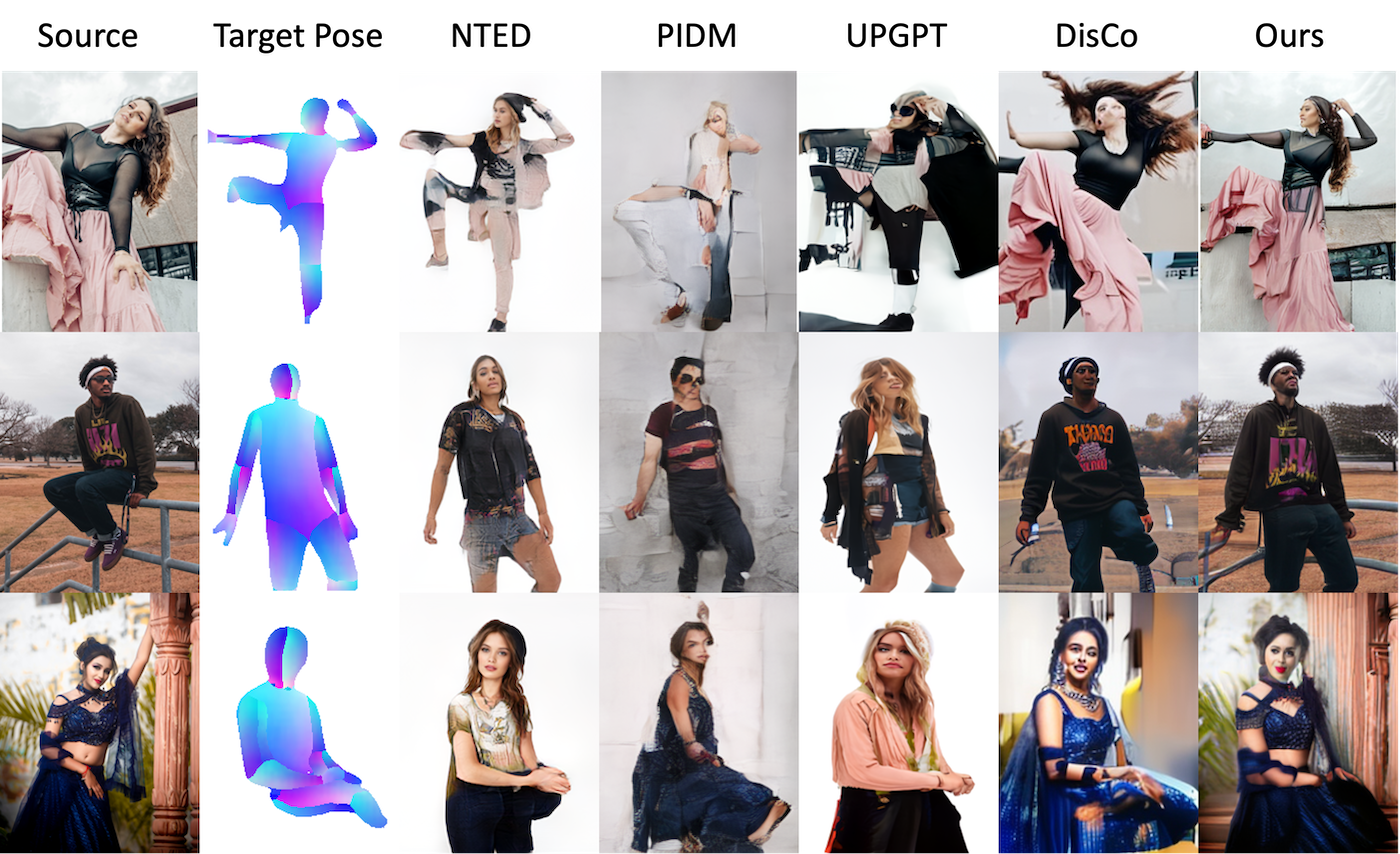}       
        \caption{Out-of-domain examples on WPose.}
        \label{fig:ood_reposing}
    \end{subfigure}
    \vspace{-0.3cm}
    \caption{Visualized results of reposing (256x256). Our model transfers the texture patterns better, particularly in out-of-domain samples. More results can be found in Supp.}
    \label{fig:reposing}
    \vspace{-0.2cm}
\end{figure*}

\begin{table*}[!t]
    \centering
    \small
    \setlength{\tabcolsep}{4pt}
    \begin{tabular}{@{}clccccccccccccc@{}}
        \toprule        
        & \multicolumn{6}{c}{In-Domain} 
        & \multicolumn{6}{c}{Out-of-Domain} \cr 
        & \multicolumn{6}{c}{DeepFashion \citep{jiang2022text2human} } 
        & \multicolumn{6}{c}{WPose} \cr 
           &\multicolumn{3}{c}{256x256} &\multicolumn{3}{c}{512x512}
         &\multicolumn{3}{c}{256x256} &\multicolumn{3}{c}{512x512} \cr
         \cmidrule(r){2-4} 
         \cmidrule(lr){5-7} 
         \cmidrule(lr){8-10} \cmidrule(l){11-13}  
        & FID$\downarrow$  & SSIM$\uparrow$ & LPIPS$\downarrow$ 
        & FID$\downarrow$  & SSIM$\uparrow$ & LPIPS$\downarrow$ 
        & FID$\downarrow$  & M-SSIM$\uparrow$ & M-LPIPS$\downarrow$         
        & FID$\downarrow$  & M-SSIM$\uparrow$ & M-LPIPS$\downarrow$ 
        \cr
        \cmidrule(r){2-4} 
         \cmidrule(lr){5-7} 
         \cmidrule(lr){8-10} \cmidrule(l){11-13}  
        
        NTED \citep{nted} &8.851 &0.766 &0.175  &7.685 &0.770 &0.186         &90.542 &0.774 &0.221  &95.953 &0.794 &0.200 \cr
        PIDM \citep{pidm} &6.401 &0.788 &0.150  &\textbf{5.899} &0.771 &0.178 &88.433 &0.774 &0.231 &92.439 &0.797 &0.211  \cr
        UPGPT \citep{cheong2023upgpt} &9.611 &0.762 &0.185 &9.698 &0.765 &0.184         &75.653 &0.784 &0.202   &85.798 &0.795 &0.196\cr
        DisCo \citep{wang2023disco} &9.842 &0.763 &0.184 &9.818 &0.770 &0.201  &50.948 &0.795 &0.181    &58.331 &0.813 &0.180 \cr        
        \Ours{} &\textbf{5.089} &\textbf{0.815} &\textbf{0.123}
        &6.092 &\textbf{0.807} &\textbf{0.148}
        &\textbf{27.571} &\textbf{0.810} &\textbf{0.159}
        &\textbf{27.748} &\textbf{0.824} &\textbf{0.147} \cr
        \bottomrule
    \end{tabular}
    \vspace{-0.3cm}
    \caption{Quantitative results for reposing. Our \Ours{} outperforms current methods on both in-domain and out-of-domain data.}
    \label{tab:rp}
    \vspace{-0.3cm}
\end{table*}

The image quality criteria for LH-400K are: 1) image resolution is no less than 512 pixels; 2) image contains only one person; 3) head is visible; 4) pose can be detected by pose detection models; 5) little to no occlusion between the human and other objects in the scene; 6) clothing covers an adequate area of the human 7) image-text similarity is larger than 0.2. To fulfill these criteria, we applied human detector \citep{ren2015faster}, face detector \citep{mullapudi2018hydranets}, pose detector \citep{openpose,guler2018densepose}, instance segmentation model \citep{cheng2022masked}, human parser \citep{parser} and CLIP \citep{clip} on images from LAION-400M. Our resulting dataset contains 409,270 clean human images with diverse backgrounds.

\subsection{Test Data Collection}
\label{sec:test data}
To evaluate the model's generalization capability to real-world unseen data, we collect two out-of-domain test sets: WPose for human reposing; and WVTON for virtual try-on. WPose comprises 2,304 real-world human image pairs with individuals adopting diverse postures (\ie, dancing, squatting, lying down, \textit{etc}.) that are rarely encountered in the training data. The backgrounds in WPose also encompass a diverse spectrum, ranging from indoor settings to outdoor scenes. Similarly, in WVTON, we collected 440 test pairs using garment images from Stock photos that comprise clothing items with diverse graphic patterns and fabric textures, to serve as our in-the-wild test set for virtual try-on. We confirmed our test data have a CC-0 or comparable license that allows academic use.

\section{Experiments}

\paragraph{Datasets}
See \cref{tab:data} for dataset statistics. We train one model each for 256x256 and 512x512 images. As a unified model, \textbf{one} single trained model is used to evaluate on \textbf{all} tasks and on both in-domain and out-of-domain samples. See Supp. for our implementation details.

\smallskip
\paragraph{Metrics} 
We use Structural Similarity Index Measure (SSIM) \citep{ssim}, Frechet Inception Distance (FID) \citep{fid}, Kernel Inception Distance (KID) \citep{kid}, and Learned Perceptual Image Patch Similarity (LPIPS) \citep{lpips} to evaluate image quality. On in-the-wild reposing test data, we use Masked-SSIM and Masked-LPIPS computed on the target person region to exclude the irrelevant background from contributing to the metrics. All the compared methods use the same image size and padding when computing the above metrics.

\begin{table}[t]
    \centering
    \small
    \setlength{\tabcolsep}{1.6pt}
    \begin{tabular}{@{}lcccc@{}}
    \toprule
     Methods &PIDM \citep{pidm} &\Ours{} &DisCo \citep{wang2023disco} &\Ours{} \cr \cmidrule(r){2-3} \cmidrule(l){4-5} 
     Pose Accuracy &15.7\%  &\textbf{84.3\%} &27.9\% &\textbf{72.1\%}\cr
     Texture Consistency &12.4\% &\textbf{87.6\%} &35.9\% &\textbf{64.0\%} \cr 
     Face Identity &11.5\%  &\textbf{88.5\%}  &24.3\%  &\textbf{75.7\%}  \cr
    \bottomrule
    \end{tabular}
    \vspace{-0.2cm}
    \caption{Human evaluation results on WPose for reposing. Our \Ours{} is preferred by users on all three evaluation methods.  }
    \label{tab:heval_reposing}
    \vspace{-0.2cm}
\end{table} 
\begin{figure*}[!t]
    \centering
    \begin{subfigure}[t]{0.384\textwidth}
        \centering
        \includegraphics[width=\textwidth]{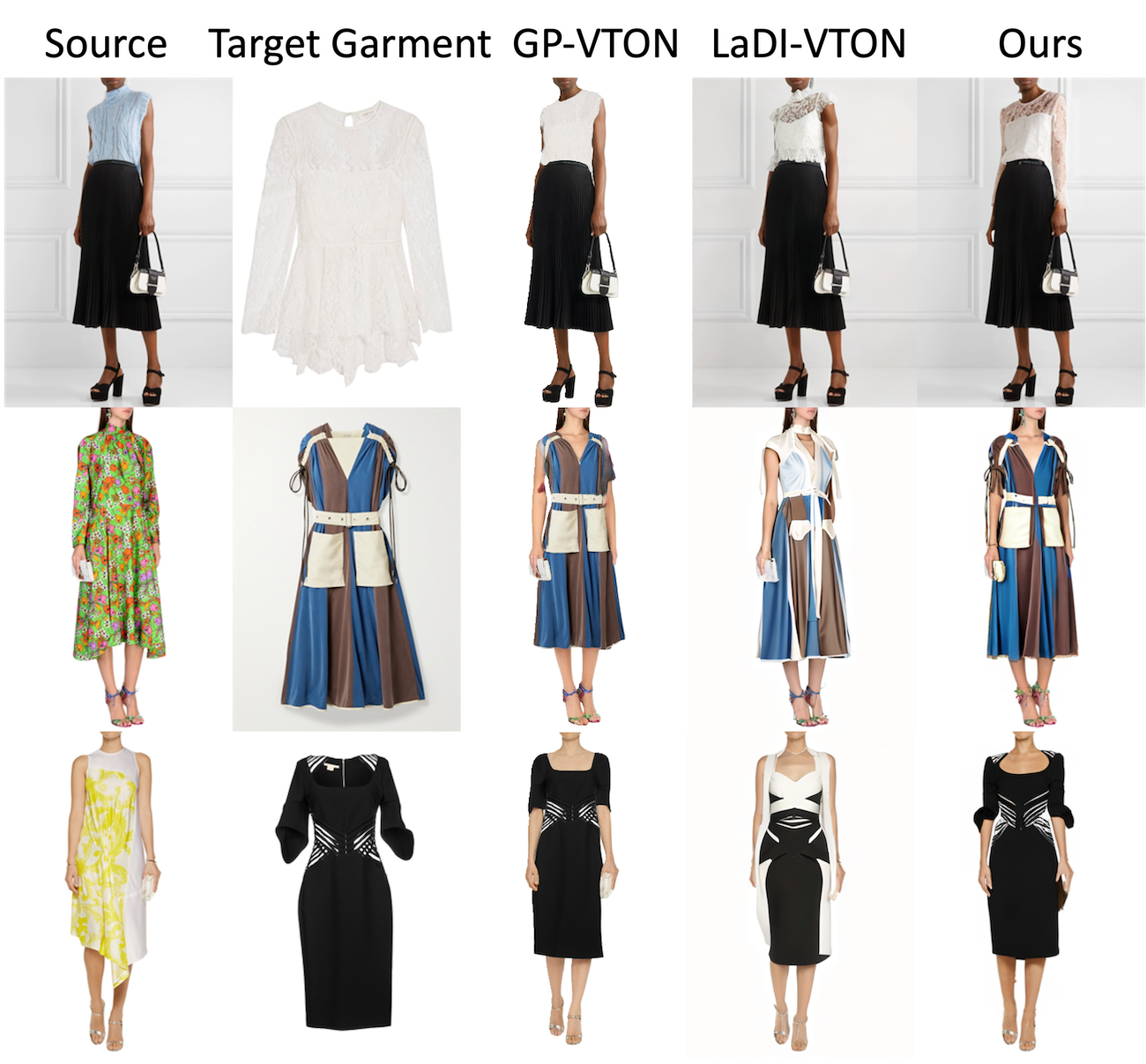}        
        \caption{In-domain examples on DressCode \citep{dresscode}.}
        \label{fig:iid_tryon}
    \end{subfigure}
    \rulesep
    \begin{subfigure}[t]{0.596\textwidth}
        \centering
        \includegraphics[width=\textwidth]{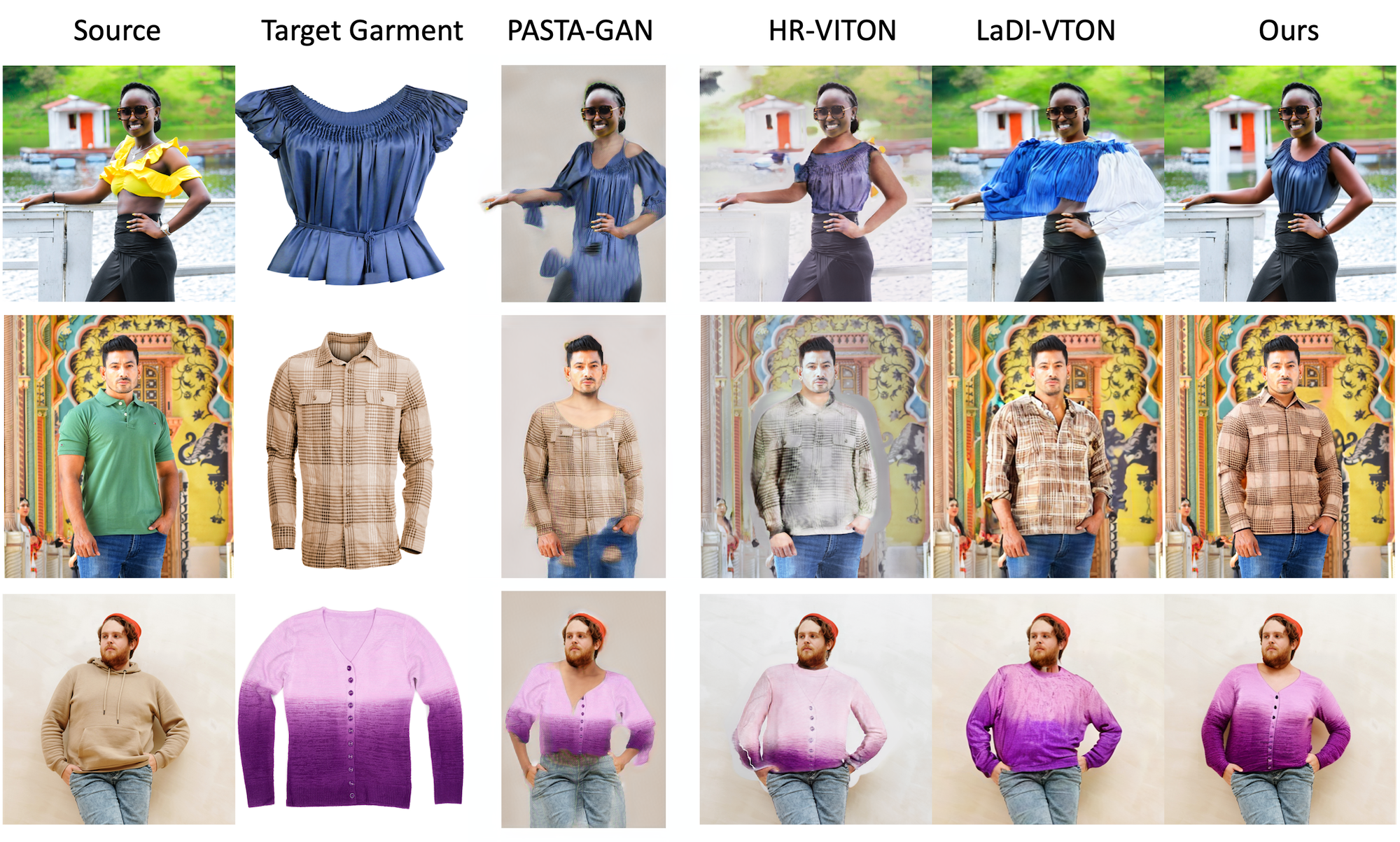}       
        \caption{Out-of-domain examples on WVTON.}
        \label{fig:ood_tryon}
    \end{subfigure}
    \vspace{-0.2cm}
    \caption{Virtual try-on results (512 x 512). Our \Ours{} better recovers the intricate details in the target garment, particularly in out-of-domain samples. More results can be found in Supp.}
    \label{fig:tryon}
    \vspace{-0.1cm}
\end{figure*}

\begin{table*}[!t]
    \centering
    \small    
    \setlength{\tabcolsep}{3.5pt}
    \begin{tabular}{@{}lcccccccccccccc@{}}
        \toprule
        & \multicolumn{12}{c}{In-Domain} & \multicolumn{2}{c}{Out-of-Domain} \cr 
         & \multicolumn{6}{c}{VITON-HD \citep{vitonhd}}  & \multicolumn{6}{c}{DressCode \citep{dresscode}} 
         &\multicolumn{2}{c}{WVTON}  \cr
         \cmidrule(r){2-7} 
         \cmidrule(lr){8-13}  \cmidrule(l){14-15} 
         
          & \multicolumn{4}{c}{Paired} & \multicolumn{2}{c}{Unpaired} 
         & \multicolumn{4}{c}{Paired} & \multicolumn{2}{c}{Unpaired}
         & \multicolumn{2}{c}{Unpaired}  \cr 
         \cmidrule(r){2-5} \cmidrule(lr){6-7} 
         \cmidrule(lr){8-11} \cmidrule(lr){12-13}  \cmidrule(l){14-15} 
        & FID$\downarrow$ & KID$\downarrow$ & SSIM$\uparrow$ & LPIPS$\downarrow$ 
        & FID$\downarrow$ & KID$\downarrow$ 
        & FID$\downarrow$ & KID$\downarrow$ & SSIM$\uparrow$ & LPIPS$\downarrow$ 
        & FID$\downarrow$ & KID$\downarrow$ 
        & FID$\downarrow$ & KID$\downarrow$ \cr
        \cmidrule(r){2-5} \cmidrule(lr){6-7} 
         \cmidrule(lr){8-11} \cmidrule(lr){12-13}  \cmidrule(l){14-15} 
        PASTA-GAN \citep{pasta} &-     &-     &-     &-     &-     &-     &-      &-     &-     &-     &-       &-       &179.138 &6.561\cr
        HR-VITON \citep{hrviton} &7.533 &0.160 &0.902 &0.075 &9.979 &0.200  &-      &-     &-     &-      &-      &-        &159.553 &3.269\cr
        GP-VTON \citep{xie2023gp} &5.572 &0.069 &\textbf{0.913} &\textbf{0.064} &8.641 &0.589 
        &9.111 &0.584 &0.900 &0.080 &10.600 &1.205     &-   &-\cr
        LaDI-VTON \citep{ladiviton} &5.647 &0.047 &0.901 &0.070 &\textbf{8.249} &0.078 &4.820 &0.106 &0.920 &0.059 &6.727 &0.182  &147.375 &2.416\cr        
        \Ours{} &\textbf{5.238} &\textbf{0.036} &0.905 &0.068 
        &8.312 &\textbf{0.072}
        &\textbf{3.947} &\textbf{0.059} &\textbf{0.929} &\textbf{0.053} &\textbf{5.529} &\textbf{0.134}
        &\textbf{127.866} &\textbf{1.671}  \cr
       
        \midrule
        GP-VTON(w.o. bg) &\multicolumn{6}{c}{----} &5.649 &0.234 &0.930 &0.043 &7.067 &0.311 &\multicolumn{2}{c}{-}\cr
        \Ours{}(w.o. bg) &\multicolumn{6}{c}{----} &\textbf{3.446} &\textbf{0.067} &\textbf{0.935} &\textbf{0.040} &\textbf{5.558} &\textbf{0.139} &\multicolumn{2}{c}{-}\cr
        \bottomrule
    \end{tabular}
    \vspace{-0.2cm}
    \caption{Quantitative results for virtual try-on (512x512). KID is multiplied by 100. Our \Ours{}, outperforms current methods on most metrics, particularly in out-of-domain data.}
    \label{tab:vt}
\end{table*}
\begin{table}[!t]
    \centering
    \setlength{\tabcolsep}{0.7pt}
    \scalebox{0.85}{
    \begin{tabular}{@{}lcccc@{}}
    \toprule
    Methods &HR-VITON &\Ours{} &LaDI-VTON &\Ours{} \cr \cmidrule(r){2-3} \cmidrule(l){4-5} 
     Texture Consistency &23.6\% &\textbf{76.4\%} &24.1\% &\textbf{75.9\%} \cr 
     Image Fidelity &21.6\%  &\textbf{78.4\%}  &27.7\%  &\textbf{72.3\%}  \cr
    \bottomrule
    \end{tabular}
    }
    \vspace{-0.2cm}
    \caption{User studies on out-of-domain test data for the virtual try-on task. HR-VITON \citep{hrviton} and LaDI-VTON \citep{ladiviton} are virtual try-on methods. Our model is preferred in at least 72\% cases.}
    \label{tab:heval_tryon}
    \vspace{-0.1cm}
\end{table}

\subsection{Reposing Experiments}
\label{sec:reposing}
\paragraph{Quantitative Results}
\cref{tab:rp} compares our methods with NTED\citep{nted}, PIDM \citep{pidm} and UPGPT \citep{cheong2023upgpt}, DisCo \citep{wang2023disco}. NTED and PIDM are reposing methods trained on DeepFashion. UPGPT is a multi-task model trained on DeepFashion. DisCo is a reposing approach trained on combined datasets of 700K images, including DeepFashion, TiKTok dance videos \citep{tiktok}, and \emph{etc}. Since DisCo has not yet released its model at 512x512 resolution, we upsampled its generated 256x256 images and reported the numbers. In \cref{tab:rp}, PIDM has a slightly better FID score at 512x512 resolution on DeepFashion but worse scores on other metrics, suggesting that it generates more realistic texture yet fails to keep the identities of the original clothing. Overall, our model shows significant gains on most metrics, particularly on out-of-domain test data. Our model boosts the FID from $58$ to $27$ at 512x512 image resolution, demonstrating the better generalization capacity of the proposed method.
\smallskip

\paragraph{Qualitative Results}
\cref{fig:iid_reposing} shows our model's capacity at reconstructing intricate details. For example, note the equally distanced stripes in the first row and the accurate preservation of triangle patterns in the second row. For out-of-domain test samples in \cref{fig:ood_reposing}, models trained on a single dataset (PIDM, NTED, and UPGPT) exhibit a recurring limitation. They transform the backgrounds, faces, and clothing to be similar to training samples, indicating poor generalization. Even though DisCo utilized more paired images from a video dataset to train their model, the last two rows of \cref{fig:ood_reposing} show the resulting images have notable texture change after reposing. In contrast, our \Ours{} preserves the clothing identity in synthesized images, \eg, the graphic logo on the black t-shirt in the middle row.
\smallskip

\begin{table*}[!t]
    \centering
    \small
    \setlength{\tabcolsep}{3pt}
    \begin{tabular}{@{}llcccccccccccc@{}}
        \toprule
         && \multicolumn{7}{c}{In-Domain} & \multicolumn{5}{c}{Out-of-Domain} \cr 
         && \multicolumn{3}{c}{DeepFashion \citep{jiang2022text2human}} & \multicolumn{2}{c}{VITON-HD \citep{vitonhd}} 
         & \multicolumn{2}{c}{DressCode \citep{dresscode}} & \multicolumn{3}{c}{WPose} 
         & \multicolumn{2}{c}{WVTON} \cr
         
         \cmidrule(lr){3-5} \cmidrule(lr){6-7} 
         \cmidrule(lr){8-9} \cmidrule(lr){10-12}  \cmidrule(l){13-14} 
        & & FID$\downarrow$ & SSIM$\uparrow$ & LPIPS$\downarrow$ 
        & FID$\downarrow$ & KID$\downarrow$ & FID$\downarrow$ & KID$\downarrow$ 
        & FID$\downarrow$ & M-SSIM$\uparrow$ 
        & M-LPIPS$\downarrow$ & FID$\downarrow$ & KID$\downarrow$ \cr
       
         \cmidrule(lr){3-5} \cmidrule(lr){6-7} 
         \cmidrule(lr){8-9} \cmidrule(lr){10-12}  \cmidrule(l){13-14} 

        \multirow{2}{*}{(a)}
        &w.o. $I_{tex}$ &6.016 &0.798 &0.143 
        &9.953 &0.295 &7.042 &0.279 
        &31.352 &0.797 &0.178
        &146.352 &2.134\cr
        
        &w.o. $L_E$ &5.232 &0.802 &0.131  
        &9.584 &0.185 &6.981 &0.275
        &30.267 &0.802 &0.169  
        &140.187 &2.036\cr

        \midrule

        \multirow{3}{*}{(b)}
        &w.o. 400K \& $I_{tex}$ 
        &6.155 &0.813 &0.123  
        &9.998 &0.316 &7.210 &0.280 
        &47.736 &0.791 &0.178 
        &165.736 &3.120\cr        
        
        &w.o. 400K &5.269 &\textbf{0.826} &\textbf{0.110} 
        &9.659 &\textbf{0.184}  &6.960 &0.271 
        &37.826 &0.804 &0.167 
        &142.839 &2.139\cr

        &rp only &5.682 &0.813 &0.124 
        &17.345 &1.005 &14.909 &0.867
        &42.587 &0.790 &0.185
        &170.322 &3.180\cr
        
        \midrule
        
        &\Ours{}  &\textbf{5.089} &0.815 &0.123 
        &\textbf{9.558} &0.248 &\textbf{6.310} &\textbf{0.208} 
        &\textbf{27.571} &\textbf{0.810} &\textbf{0.159} 
        &\textbf{131.500} &\textbf{1.730}\cr
        \bottomrule
    \end{tabular}
    \vspace{-0.2cm}
    \caption{Ablation results. Image resolution is 256x256. KID is multiplied by 100. Our full model achieves the best performance.}
    \label{tab:abl}
    \vspace{-0.5cm}
\end{table*}
\begin{figure}[!t]
    \centering
    \includegraphics[width=0.49\textwidth]{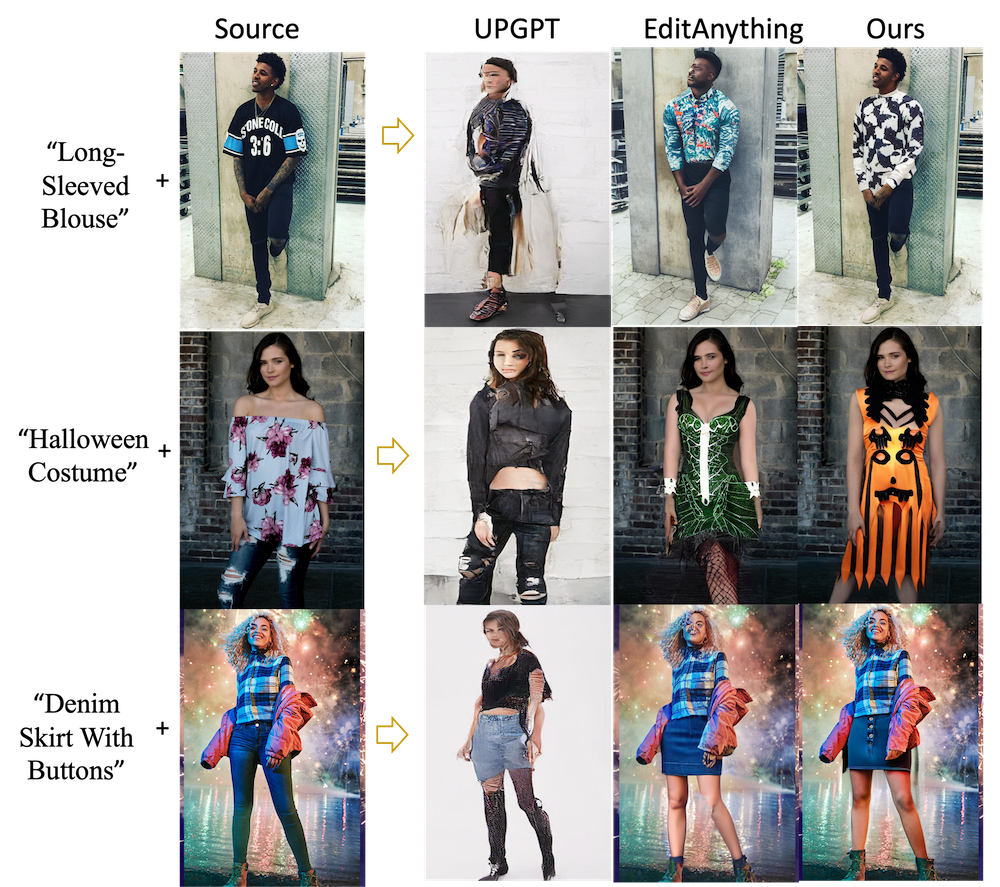}
    \vspace{-0.5cm}
    \caption{Results of text manipulation. Our model manipulates the clothing to match the specified concept.}
    \label{fig:text_manip}
    \vspace{-0.3cm}
\end{figure}

\paragraph{User Studies}
We randomly chose 200 images from the out-of-domain test data WPose, and each generated image was evaluated by three workers from Amazon MTurk to compare with the synthesized image of an existing method (\ie, PIDM or DisCo). These workers evaluate the image quality on three aspects: pose accuracy, clothing texture consistency, and face identity consistency.  \cref{tab:heval_reposing} reports the study results, where our method is preferred in all three aspects.

\subsection{Virtual Try-on Experiments}
\paragraph{Results}
\cref{tab:vt} reports our results with state-of-the-art performance over other methods specifically designed for virtual try-on.  Note that the original GP-VTON \citep{xie2023gp} removed all backgrounds in DressCode, and in this setting, we still outperform them (the last two lines of \cref{tab:vt}). Some of our gains are likely attributed to the reposing data we can take advantage of due to using a unified model.  Specifically, there are only 120,078 virtual try-on data samples in our training set, but 453,366 reposing samples. Additionally, compared with try-on-only models, we also generalize better to out-of-domain samples, where our FID score improves from $147$ to $128$. As shown in \cref{fig:tryon}, our model successfully reconstructs the texture patterns of unseen garments for a variety of body shapes.
\smallskip

\paragraph{User Studies}
Similar to the human evaluation in \cref{sec:reposing}, we conducted user studies for virtual try-on. This time, we asked the workers to evaluate the image fidelity and the texture consistency of the try-on garment, respectively. \cref{tab:heval_tryon} shows that our model is preferred in at least 72\% of test cases, demonstrating the superior performance of the proposed method for in-the-wild scenarios.

\subsection{Text Manipulation Experiments}
The proposed method can manipulate the clothing shape, category, and texture by text description, as showcased in \cref{fig:text_manip}. We chose UPGPT \citep{cheong2023upgpt} and EditAnything \citep{gao2023editanything} as our baselines since both methods have been used for fashion item editing. Notably, our model manages to maintain the original pose while manipulating the image content to match the specified concept. For example, both EditAnything and our model turn the lady's clothing into a Halloween costume on the second row, whereas only our method keeps the original body pose unchanged. See Supp. for more analysis about text manipulation.


\subsection{Ablations}



\paragraph{Pose-Warped Texture}
In \cref{tab:abl}(a), \textbf{w.o. \bm{$I_{tex}$}} removes the pose-warped texture. Consequently, we notice a marked drop in performance across all metrics, underscoring the critical role played by the pose-warped texture and the significance of visible pixels it provides. In a separate experiment, \textbf{w.o. \bm{$L_E$}} includes the pose warped-texture but removes the loss function $L_E$, which constrains the attention area of the pose-warped texture. This model shows better performance than \textbf{w.o. \bm{$I_{tex}$}}, but it still has a significant gap with our full model. By introducing the pose-warping module and $L_E$, our full model yields the most substantial gains across both in-domain and out-of-domain test sets.

\smallskip
\paragraph{Data}
To validate the effectiveness of the collected LH-400K, we train an ablation model with only existing datasets, denoted by
\textbf{w.o. 400K} in \cref{tab:abl}(b). Removing LH-400K results in comparable or even superior in-domain performance (\ie, better SSIM and LPIPS). However, out-of-domain metrics suffer from a significant drop, indicating overfitting and poor generalization due to limited data scale and diversity. In \textbf{rp only}, we exclude both LH-400K and try-on datasets, relying solely on DeepFashion for training the reposing task. The performance on both reposing datasets, \ie, in-domain DeepFashion and out-of-domain WPose, has a notable decline compared to \textbf{w.o. 400K}. This demonstrates the beneficial impact of the try-on data on the reposing task. Additionally, 
further removing $I_{tex}$ (\textbf{w.o. 400K \& \bm{$I_{tex}$}}) causes a more pronounced drop, emphasizing the importance of the pose-warped texture.
More ablation experiments can be found in Supp.

\section{Conclusion}
We propose \Ours{}, a unified model that achieves high-quality human image editing results on multiple tasks for both in-domain and out-of-domain data. We leverage human visual encoders and a lightweight pose-warping module that exploits different pose representations to accommodate unseen textures and patterns. Furthermore, to expand the diversity and scale of existing datasets, we curated 400K high-quality image-text pairs for training and collected 2K human image pairs for out-of-domain testing. Experiments on both in-domain and out-of-domain test sets demonstrate that \Ours{} outperforms task-specific models by a significant margin, earning user preferences in an average of 77\% of cases. In future work, we will explore the adaptation of \Ours{} to the video domain.

{
    \small
    \bibliographystyle{ieeenat_fullname}
    \bibliography{myBib}
}

\clearpage
\setcounter{page}{1}
\maketitlesupplementary


In the following, we first discuss related literature in \cref{sec:more related work}. Then we analyze our text manipulation results in \cref{sec:sup_tm} and show more ablation experiments in \cref{sec:more_abl}. Subsequently, we explain how to incorporate LH-400K dataset into training in \cref{sec:unpaired train} and show that our unified model can achieve multi-task combinations in \cref{sec:multi-task}. Finally, we explain our implementation details in \cref{{sec:imp details}} and present various visualized examples in \cref{sec:more vis}.

\section{Task-Specific Models Discussion}
\label{sec:more related work}

In the following, we discuss existing task-specific models that achieve reposing, virtual try-on, and text manipulation in the 2D image domain and how they differ from our proposed approach. Since our goal is human \textit{image} editing, we do not compare with 3D and video based models \citep{xu2022surface,santesteban2022ulnef,3d-LiuPTLMG22,3d-WangCSM21,dong2022dressing} as they are out-of-scope of this paper.

\paragraph{Reposing}
To change a person's pose, previous approaches typically encode the person as a whole and learn the transformation across poses \citep{pidm,cheong2023upgpt,wang2023disco,dreampose,zhang2022exploring,ren2021combining}. Adapting these models to handle multi-task scenarios where only specific body parts need to be modified can be difficult.
Other methods enhance the versatility of reposing models by dissecting the person into different parts \citep{Cui_2021_ICCV,ma2021must,wang2022self,li2023collecting,cheong2023upgpt}, thus allowing independent editing of each part. However,  relying solely on part-wise texture information may lead to challenges in recognizing the identities of individual body parts. For example, a strapless top might be misidentified as a mini skirt since both clothing could have similar textures and shapes. To address this concern, a word embedding labeling the clothing type, such as \textit{upper clothing} and \textit{lower clothing}, is concatenated with the DINOv2 features. Furthermore, we apply a loss $L_B$ in \cref{sec:loss} to localize the cross-attention map of each human part to its corresponding region. We find these strategies effectively improve the performance of our model. 

For reposing methods that introduce a pose-warping module \citep{grigorev2019coordinate,albahar2021pose}, a UV coordinate inpainting model was trained to infer invisible pixels from their visible counterparts, which is unsuitable for warping in-shop garments that lack such UV representation. As a unified model, \Ours{} can utilize both dense pose UV representation and sparse keypoint locations to warp clothing texture to the RGB space, ensuring the provision of accurate visible pixel information across domains.

\smallskip
\paragraph{Virtual Try-on}
The objective of virtual try-on is to seamlessly fit the target in-shop clothing to a person \citep{pasta,ladiviton,zhu2023tryondiffusion,hrviton,vitonhd}. 
In prior work, this is often accomplished through a two-stage process where the clothing is initially warped through a deep learning model and subsequently aligned with the person in a second model \citep{li2023virtual,xie2023gp,hrviton,vitonhd,yang2020towards,dong2022dressing}. 
The clothes warping module often learns the parameters of a Thin-Plate Spline transformation (TPS) \citep{duchon1977splines} from the target garment to the target pose \citep{yang2020towards}. To balance the flexibility of TPS with the rigidity of affine transformation, researchers have introduced various regularization terms to train these parameters \citep{ge2021disentangled,yang2020towards,yang2022full}.
However, these learned warping modules are trained using try-on data to establish correspondences between the clothing and the pose, posing a risk of overfitting to specific body shapes within the dataset. In contrast, our pose-warping module harnesses the pose correspondences to map the original pixels to pose-warped texture without training. Moreover, while TPS is not suitable for the pose transfer task due to the non-smoothness of the pose transformation, our pose-warping model can leverage dense pose for texture warping.

\smallskip
\paragraph{Diffusion Based Text Manipulation}
The advent of diffusion models has ushered in a new era in image editing through text descriptions \citep{chang2023muse,hertz2022prompt,goel2023pair,yang2023paint,nichol2022glide}. Among these methods, latent Stable Diffusion (SD) \citep{stablediff} has gained popularity because of its versatility in accommodating prompts of various formats, coupled with its efficient memory utilization within the latent space. However, the challenge of editing human images using text prompts persists, primarily due to the highly structured nature of the human body \citep{jiang2022text2human,gao2023editanything,cheong2023upgpt}. Additionally, enhancing the alignment between text and images requires the image captions to include information about clothing categories, shapes, and textures. In our pursuit of expanding existing human image-text datasets \citep{jiang2022text2human}, we curated a new dataset featuring single human images paired with captions from LAION-400M \citep{schuhmann2021laion400m}. We believe that incorporating these image-text pairs in the human image editing tasks can further improve the data diversity and enrich the modality of our model.

\begin{figure*}[!t]
    \centering
    \includegraphics[width=0.97\textwidth]{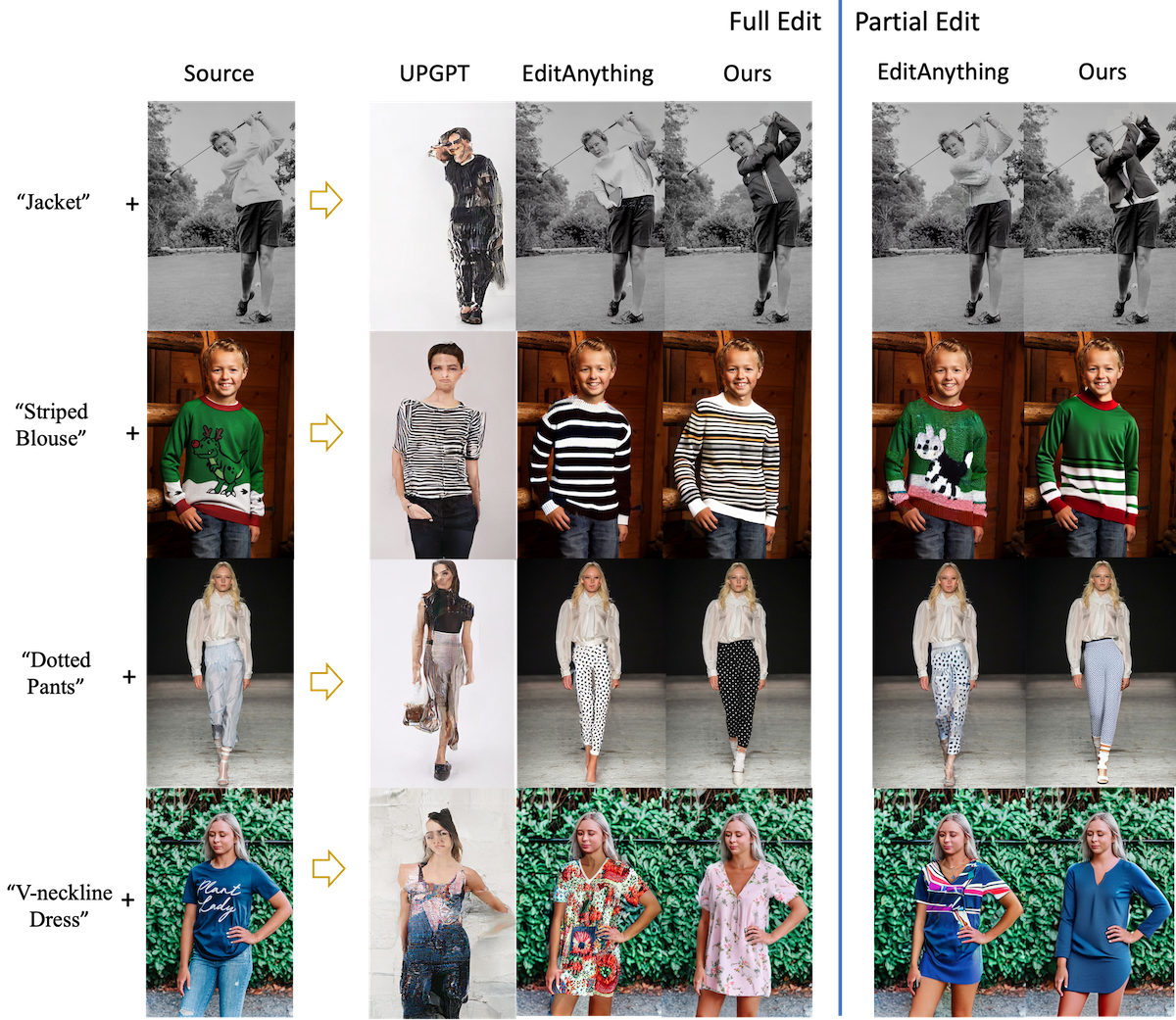}
    \vspace{-0.2cm}
    \caption{Text manipulation examples. Full Edit indicates a new garment is synthesized from scratch. Partial Edit means partial textures from the source garment are used to generate the new clothing. Zoom in to see details.}
    \label{fig:sup_tm}
    \vspace{-0.2cm}
\end{figure*}

\begin{table*}
\centering
\begin{minipage}[t]{0.35\textwidth}
     \centering
    \small
    \begin{tabular}{@{}lccc@{}}
    \toprule
    &FID$\downarrow$ &KID$\downarrow$ &CLIP$\uparrow$ \cr
    \cmidrule{2-4}
    UPGPT \citep{cheong2023upgpt} &138.178 &8.548 &13.628\cr
    EditAnything \citep{gao2023editanything} &63.336 &1.986 &15.235\cr
    \Ours{} &\textbf{62.663} &\textbf{1.782} &\textbf{15.715}\cr
    \bottomrule
    \end{tabular}
    \vspace{-0.1cm}
    \captionof{table}{Quantitative results for text manipulation. Our \Ours{} shows outperforms the baselines in all metrics.}
    \label{tab:quan_tm}
\end{minipage}
\vspace{-0.2cm}
\hfill
\begin{minipage}[t]{0.6\textwidth}
    \centering
    \small
    \setlength{\tabcolsep}{2.5pt}
    \begin{tabular}{@{}lcccc@{}}
    \toprule
     Methods &EditAnything \citep{gao2023editanything}  &\Ours{} &UPGPT \citep{cheong2023upgpt} &\Ours{} \cr \cmidrule(r){2-3} \cmidrule(l){4-5} 
     Pose Accuracy &32.6\% &67.4\%  &13.2\% &86.8\% \cr
     Image Plausibility &35.8\%  &64.2\% &15.9\%  &84.1\% \cr 
     Image-Text Similarity &39.5\% &60.5\% &17.8\% &82.2\% \cr
    \bottomrule
    \end{tabular}
    \vspace{-0.1cm}
    \captionof{table}{Human evaluation results on WVTON for text manipulation. Our \Ours{} is preferred by users on all three evaluation methods.  }
    \label{tab:heval_tm}
\end{minipage}
\vspace{-0.2cm}
\end{table*}

\begin{table*}[!t]
    \centering
    \small
    \setlength{\tabcolsep}{2pt}
    \begin{tabular}{@{}llcccccccccccc@{}}
        \toprule
         && \multicolumn{7}{c}{In-Domain} & \multicolumn{5}{c}{Out-of-Domain} \cr 
         && \multicolumn{3}{c}{DeepFashion \citep{jiang2022text2human}} & \multicolumn{2}{c}{VITON-HD \citep{vitonhd}} 
         & \multicolumn{2}{c}{DressCode \citep{dresscode}} & \multicolumn{3}{c}{WPose} 
         & \multicolumn{2}{c}{WVTON} \cr
         
         \cmidrule(lr){3-5} \cmidrule(lr){6-7} 
         \cmidrule(lr){8-9} \cmidrule(lr){10-12}  \cmidrule(l){13-14} 
        & & FID$\downarrow$ & SSIM$\uparrow$ & LPIPS$\downarrow$ 
        & FID$\downarrow$ & KID$\downarrow$ & FID$\downarrow$ & KID$\downarrow$ 
        & FID$\downarrow$ & M-SSIM$\uparrow$ 
        & M-LPIPS$\downarrow$ & FID$\downarrow$ & KID$\downarrow$ \cr
       
         \cmidrule(lr){3-5} \cmidrule(lr){6-7} 
         \cmidrule(lr){8-9} \cmidrule(lr){10-12}  \cmidrule(l){13-14} 

        \multirow{4}{*}{(c)}

        &pix seg &5.785 &0.808 &0.126 
        &9.638 &0.269 &6.345 &0.218
        &30.421 &0.808 &0.160
        &134.575 &1.964\cr
        &w.o. emd &5.456 &0.811 &0.125 
        &9.562 &0.252 &6.433 &0.209
        &28.435 &0.808 &0.161
        &132.802 &1.740\cr        
        &w.o. $L_B$ &5.827 &0.813 &0.126
        &10.086 &0.310 &6.605 &0.239
        &29.175 &0.805 &0.162
        &134.021 &1.856\cr
        &w.o. pretrain &5.328 &0.811 &0.128  
        &9.749 &0.270 &6.664 &0.252 
        &31.287 &0.805 &0.166
        &135.287 &1.998\cr

         &\Ours{}  &\textbf{5.089} &\textbf{0.815} &\textbf{0.123} 
        &\textbf{9.558} &\textbf{0.248} &\textbf{6.310} &\textbf{0.208} 
        &\textbf{27.571} &\textbf{0.810} &\textbf{0.159} 
        &\textbf{131.500} &\textbf{1.730}\cr
        
        \midrule
        \multirow{4}{*}{(d)}        
        &rp only &5.682 &0.813 &0.124 
        &17.345 &1.005 &14.909 &0.867
        &42.587 &0.790 &0.185
        &170.322 &3.180\cr
        &vt only &18.081 &0.755 &0.210
        &9.721 &0.255 &7.012 &0.280
        &76.605 &0.784 &0.208
        &143.137 &2.378\cr

        &tm only &9.773 &0.751 &0.204 
        &20.623 &1.429 &18.365 &1.152
        &62.219 &0.789 &0.196 
        &175.602 &5.387 \cr

        &multi-task &\textbf{5.269} &\textbf{0.826} &\textbf{0.110} 
        &\textbf{9.659} &\textbf{0.184}  &\textbf{6.960} &\textbf{0.271} 
        &\textbf{37.826} &\textbf{0.804} &\textbf{0.167} 
        &\textbf{142.839} &\textbf{2.139}\cr

        \bottomrule
    \end{tabular}
    \vspace{-0.2cm}
    \caption{Ablation results on 256x256 images. KID is multiplied by 100. Our full model achieves the best overall performance.}
    \label{tab:more_abl}
    \vspace{-0.2cm}
\end{table*}

\section{Text Manipulation Analysis}
\label{sec:sup_tm}

\paragraph{Results}
We randomly chose 1000 images from the test sets for text manipulation evaluation. \cref{tab:quan_tm} reports the FID, KID, and CLIP image-text similarity scores for all the comparison methods. UPGPT \citep{cheong2023upgpt} is a multi-task model. EditAnything \citep{gao2023editanything} is an SD-based text manipulation model. Our \Ours{} shows better performance on all these metrics, demonstrating its capacity for human-specific text manipulation. For further evaluation, we also conducted a user study. We asked AMTurk workers to compare 200 samples edited by our model and by an existing method on WVTON. The workers evaluate the image quality on three aspects: pose accuracy, image plausibility, and image-text similarity. \cref{tab:heval_tm} reports the human evaluation results. Our \Ours{} outperforms prior work on all three aspects in maintaining the original pose and manipulating the texture, acknowledged by at least 60\% workers.

\paragraph{Visulizations}
\cref{fig:sup_tm} presents different ways of editing human images by text descriptions: the left side of the vertical ruler includes examples of generating a new garment from scratch, denoted by Full Edit; the right side of the vertical ruler shows visualizations of editing the garment given random partial source textures, denoted by Partial Edit. We did not compare with UPGPT \citep{cheong2023upgpt} in Partial Edit since it can not achieve partial editing by design. Compared with EditAnything \citep{gao2023editanything}, our \Ours{} shows better capability at making the edited image more plausible. For example, on the first row of Full Edit, our generated image better fits the jacket into the person playing golf. 

\section{Additional Ablations}
\label{sec:more_abl}

\paragraph{Part Encoder}
Our part encoder includes two modules: a DINOv2 image encoder for visual representation encoding and a CLIP Text encoder for semantic representation encoding. As mentioned in \cref{sec:more related work}, the CLIP text embedding labeling each body part helps identify these human parts. In \cref{tab:more_abl}(c), \textbf{w.o. emd} removes the text embedding, causing a slight drop in all evaluation metrics.
In another setting, to prove the effectiveness of the introduced $L_B$, we removed this loss function in \textbf{w.o. $L_B$}. Results show that the performance on both in-domain and out-of-domain test sets suffers from a larger drop than \textbf{w.o. emd}, indicating the importance of $L_B$ in our objective function. 
Additionally, \textbf{pix seg} extracts the part features at a pixel level to compare with our feature-level body part segmentation. The slight drop in all metrics shows that the contextual information from the feature-level segmentation helps reconstruct the clothing textures for all three tasks.

\smallskip
\paragraph{SD Pretraining}
We chose SD as our backbone for its excellence in producing text-aligned images and its suitability for multi-task learning. In \cref{tab:more_abl}(c), \textbf{w.o. pretrain} model takes 67\% more time to converge (5 days vs.\ 3 days) and shows a slight performance drop in the metrics.

\begin{figure*}
    \centering
    \includegraphics[width=\textwidth]{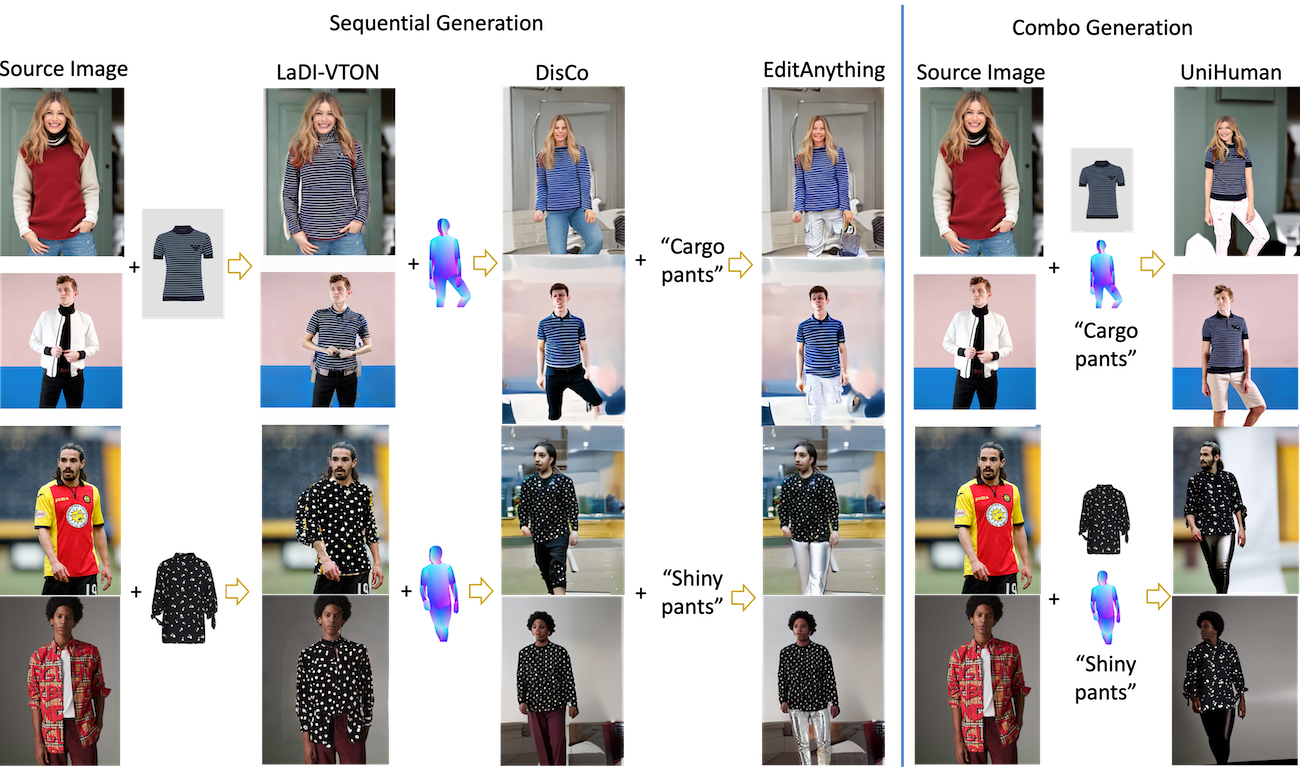}
    \vspace{-0.4cm}
    \caption{Examples of task combinations. In sequential generation, the input image sequentially goes through the virtual try-on method (LaDI-VTON), the reposing model (DisCo), and the text manipulation approach (EditAnything). In combo generation, our \Ours{} can achieve all editing tasks altogether.  }
    \label{fig:seq_vs_combo}
    \vspace{-0.2cm}
\end{figure*}

\smallskip
\paragraph{Single-Task Ablations}
To investigate if our multi-task objectives reinforce single tasks for each other, we designed four ablation models: \textbf{rp only} that takes human images from DeepFashion \citep{jiang2022text2human} to do the reposing task, \textbf{vt only} that uses human and garment images from try-on datasets \citep{dresscode,vitonhd,Minar_CPP_2020_CVPR_Workshops} to do the try-on task, \textbf{tm only} that draws image-text pairs from DeepFashion to do the text manipulation task, and \textbf{multi-task} that takes all the above data to achieve three task objectives in the same model. In \cref{tab:more_abl}(d), DeepFashion and WPose are test sets for the reposing task; VITON-HD, DressCode and WVTON are evaluation sets for the try-on task. 
\textbf{Multi-task} effectively learns all three tasks and outperforms single-task models on all metrics. This demonstrates that our multi-task objectives indeed reinforce the two visual tasks (\ie, reposing and virtual try-on) by learning them jointly. Note that \textbf{multi-task} is renamed as \textbf{w.o. 400K} in \cref{tab:abl} of the main paper.

\section{Leveraging Unpaired Images In Training}
\label{sec:unpaired train}

The visual tasks in our model require paired images for training. In reposing, we need image pairs of the same person in different poses. Recent research explores the possibility of using readily accessible unpaired human images for the reposing training \citep{ma2021must,wang2022self,li2023collecting}, which prompts us to explore the acquisition and incorporation of less costly unpaired images. Refer to \cref{sec:data} for details on collecting these data. However, the incorporation of unpaired images introduces a significant challenge. It often leads to the potential issue of pose leakage and pose-texture entanglement \citep{li2023collecting}, particularly when the volume of unpaired images surpasses that of the paired ones. To mitigate this issue, we implement strong data augmentation techniques when obtaining the part features in \cref{sec:part enc}. Beyond routine operations like image cropping, resizing and flip, we meticulously ensure that each human part is randomly warped into \textit{different orientations}. This strategy compels the model to heavily rely on the target pose to accurately restore the original body orientation, successfully addressing pose leakage and promoting a more robust and effective training process.

\section{Multi-Task Combinations in \Ours{}}
\label{sec:multi-task}
The three tasks in this paper can be combined in arbitrary ways to edit the original human image within 50 denoising time steps. Another less efficient way of accomplishing this goal is to apply these tasks sequentially using the corresponding task-specific models.
\cref{fig:seq_vs_combo} shows the results of using these two types of task combinations. Sequential Generation means applying virtual try-on, reposing, and text manipulation using Ladi-VTON, DisCo and EditAnything, sequentially on the input image. Combo Generation represents achieving all tasks simultaneously in our \Ours{}. We find that human images produced by our model better follow the target pose and the given garment textures.

\section{Implementation Details}
\label{sec:imp details}
In training, we use pretrained weights from SD v-1.5. IN the part encoder, we use ViT-B/14 for the DINOv2 visual encoding and ViT-B/16 for the CLIP text encoding. Then we fix the SD UNet encoder, the CLIP encoder, and the first 15 blocks of DINOv2, finetuning the rest layers of DINOv2 and SD UNet decoder with a learning rate of $2.5 \times 10^{-5}$. The conditioning encoder is trained from scratch with a learning rate of $10^{-4}$, which consists of three residual blocks. In our objective function, we set empirically $\lambda_1=10^{-3}$ and $\lambda_2=2.5 \times 10^{-4}$.
Both the 256-resolution model and 512-resolution model are trained for 220K iterations with a batch size of 64. In the dataloader, the target pose, pose-warped texture, part features, and the text prompt have a 10\% chance of being zero, respectively. This enables us to use classifier-free guidance \citep{ho2022classifier} when denoising the latent code, improving image quality. 

In our part-SD cross-attention, we sequentially apply a global cross-attention using the global feature representation of each human part and a local cross-attention using the patch tokens of each human part. We found this sequential attention to be slightly better than concatenating global features and patch tokens together in one attention block.  In the pose-warping module, we run DensePose \citep{guler2018densepose} pretrained on COCO to get the dense pose UV representation and use MMPose \citep{mmpose2020} pretrained on HKD \citep{wu2017ai} to obtain the sparse pose keypoint representation.


For inference, we keep the aspect ratio of the original images and resize the longer side to 256/512, then we pad the shorter side to the same size with white pixels. This type of resizing holds the original body shape to be unchanged and keeps as much background as possible. All baseline methods in our experiments use the above resizing for a fair comparison. We set $\epsilon=2$ to be the guidance scale in the classifier-free guided generation.

\begin{figure*}[!t]
    \centering
    \includegraphics[width=0.98\textwidth]{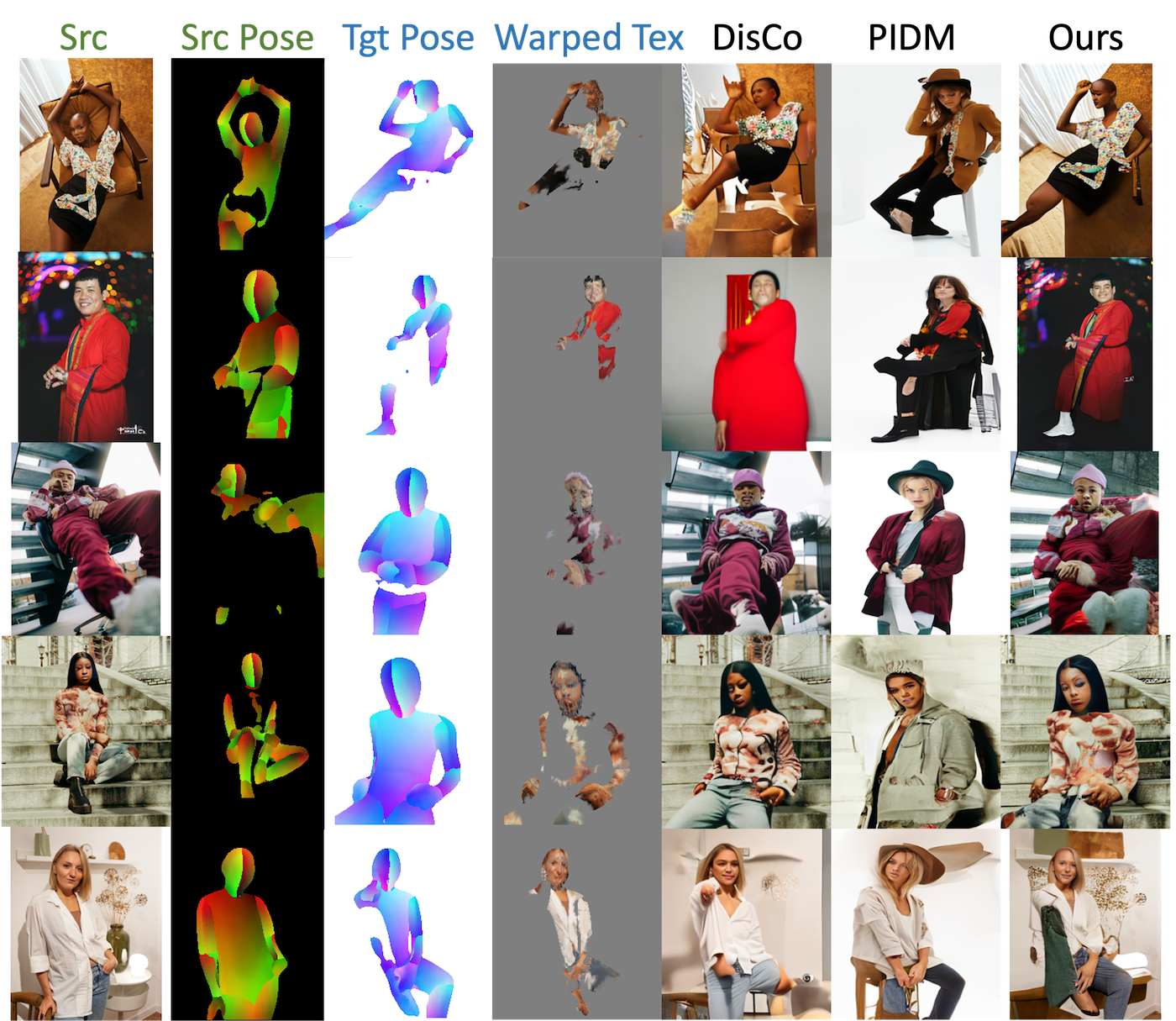}
    \caption{Examples of failed reposing generations. In cases where the source/target densepose predictions are incorrect, our model failed to transfer accurate clothing textures.}
    \label{fig:pose_fails}
\end{figure*}

In all the tasks, we paste back the background areas that do not change after editing and reconstruct the rest as in the inpainting pipeline. In the reposing task, a large position and pose change could happen, resulting in little background area in the conditioning. On the contrary, in try-on and text manipulation, most background areas do not change and thus require less inpainting.

\section{Limitation}
Our approach is limited by its reliance on pose detectors and parsing models, which is also a common limitation shared by prior work \citep{pidm,cheong2023upgpt,wang2023disco,xie2023gp}. This is an even more challenging problem in our WPose dataset, which includes diverse postures with more complicated body part occlusions than standing postures. As a result, the detected densepose could have incorrect predictions and missing parts. In \cref{fig:pose_fails}, our model failed to transfer accurate clothing textures due to densepose errors. For future work, we believe incorporating more 3D information, such as depth and surface normal, will help rectify these inaccuracies. 

\section{Additional Visualized Examples}
\label{sec:more vis}
\begin{figure*}
    \centering
    \includegraphics[width=\textwidth]{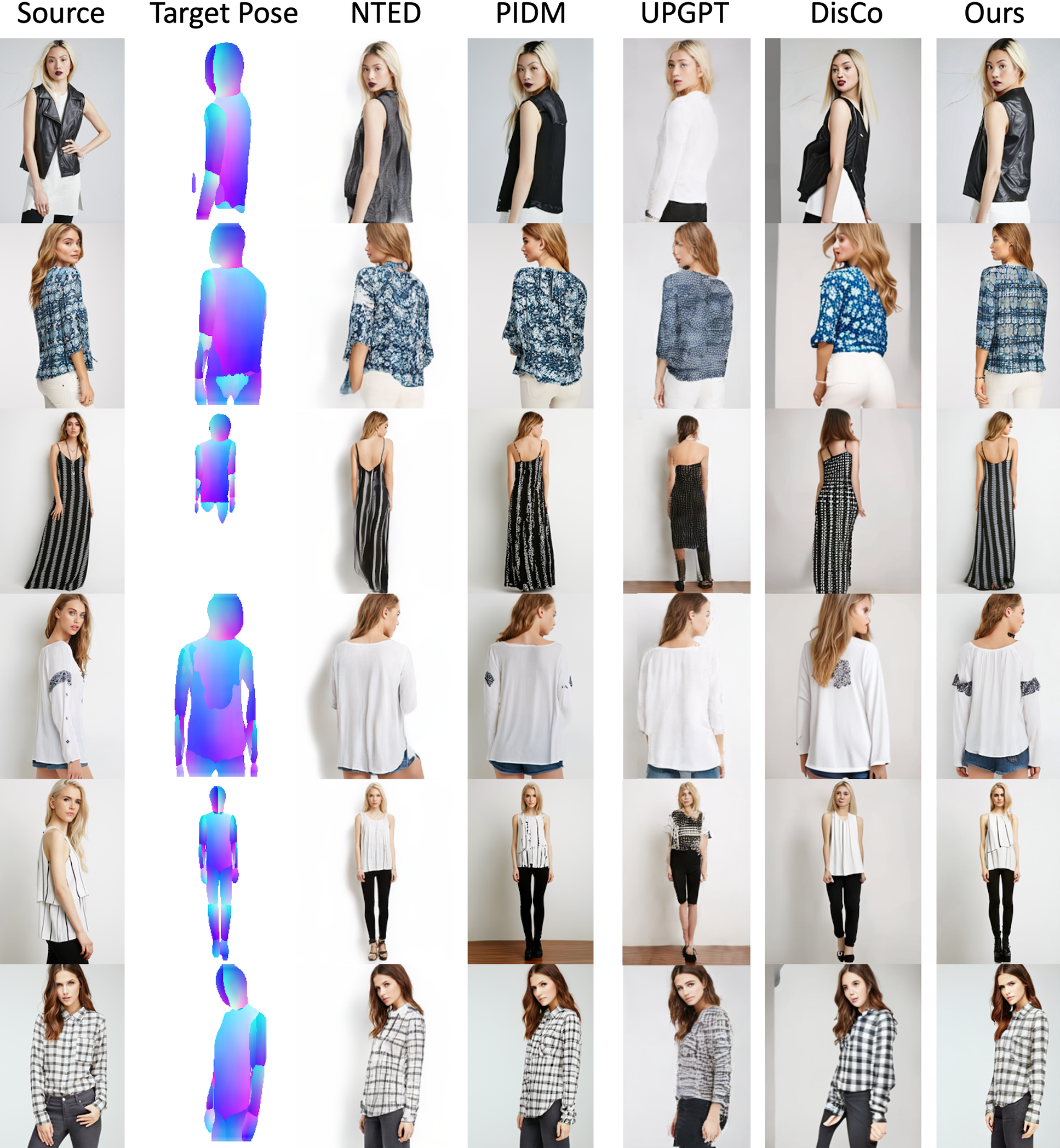}
    \vspace{-0.2cm}
    \caption{Examples of reposing on DeepFashion. Our model better reconstructs the intricate texture patterns.}
    \label{fig:more_id_repose}
    \vspace{-0.2cm}
\end{figure*}

\begin{figure*}
        \centering
    \includegraphics[width=0.97\textwidth]{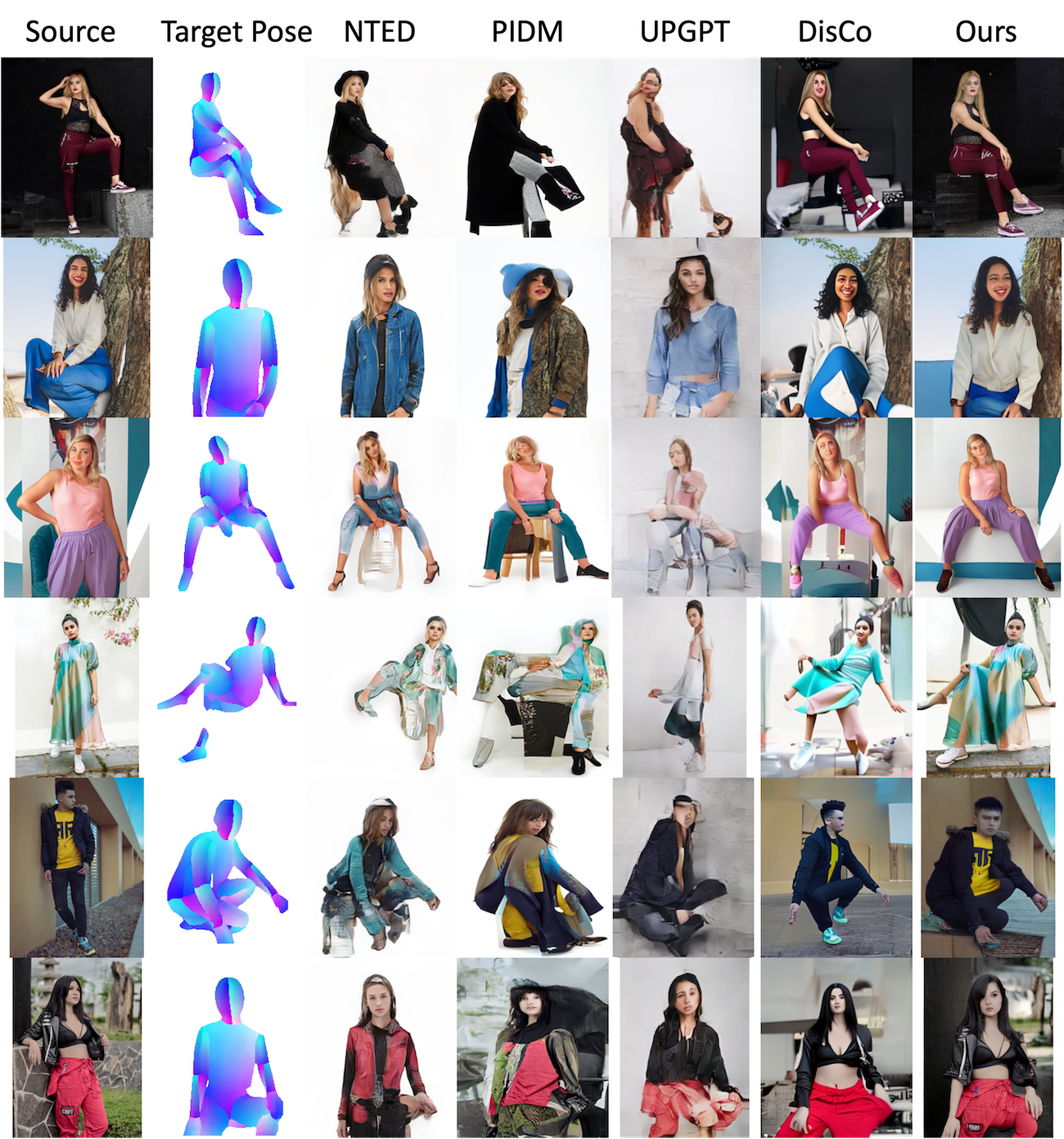}
    \caption{Examples of reposing on WPose. Images generated by our model are better aligned with the target pose while preserving the face and clothing identities.}
    \label{fig:more_ood_repose}
\end{figure*}

\begin{figure*}[!t]
    \centering
    \includegraphics[width=0.85\textwidth]{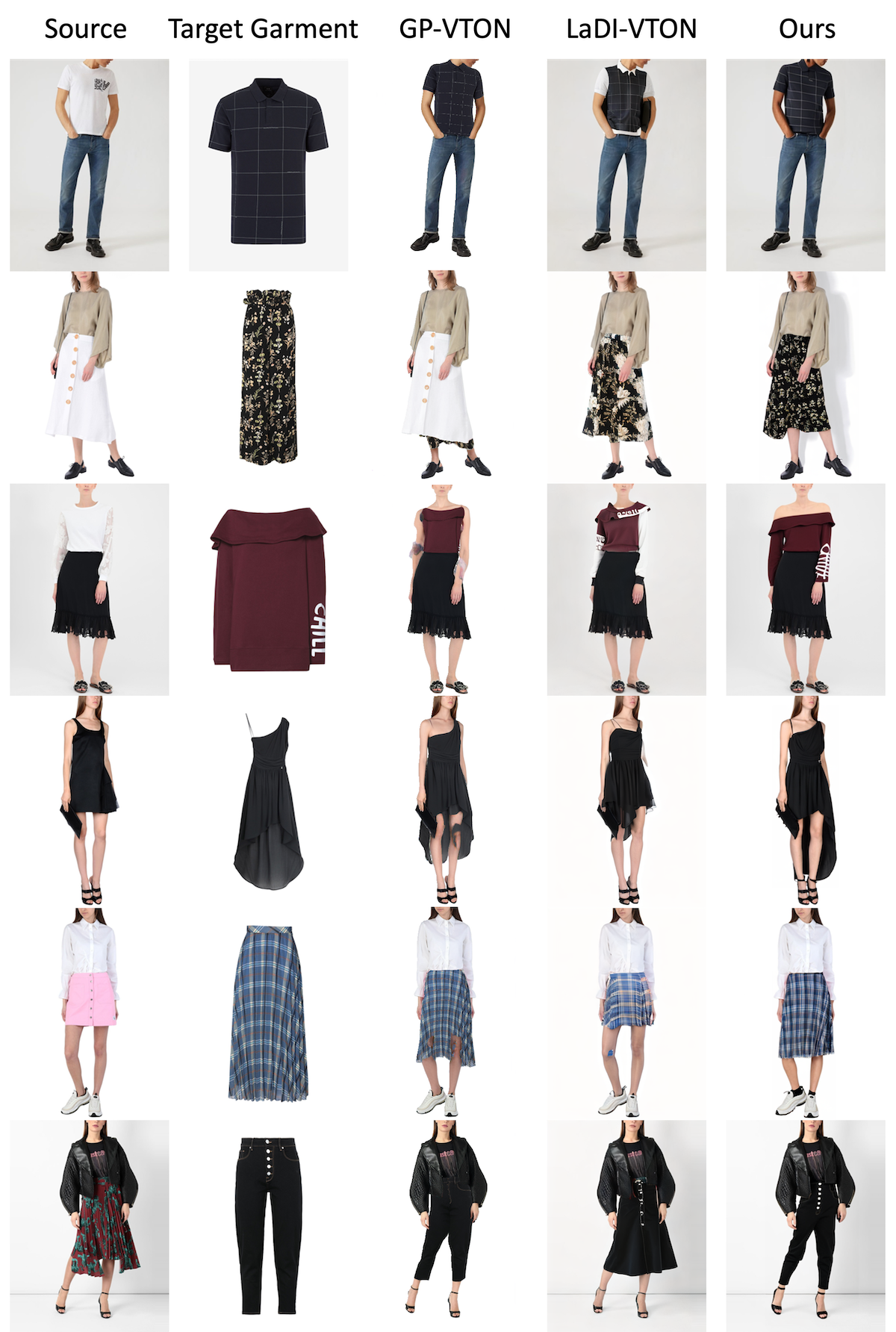}
    \caption{Examples of virtual try-on on DressCode. Our \Ours{} can recover detailed texture patterns.}
    \label{fig:more_id_tryon}
\end{figure*}

\begin{figure*}[!t]
    \centering
    \includegraphics[width=\textwidth]{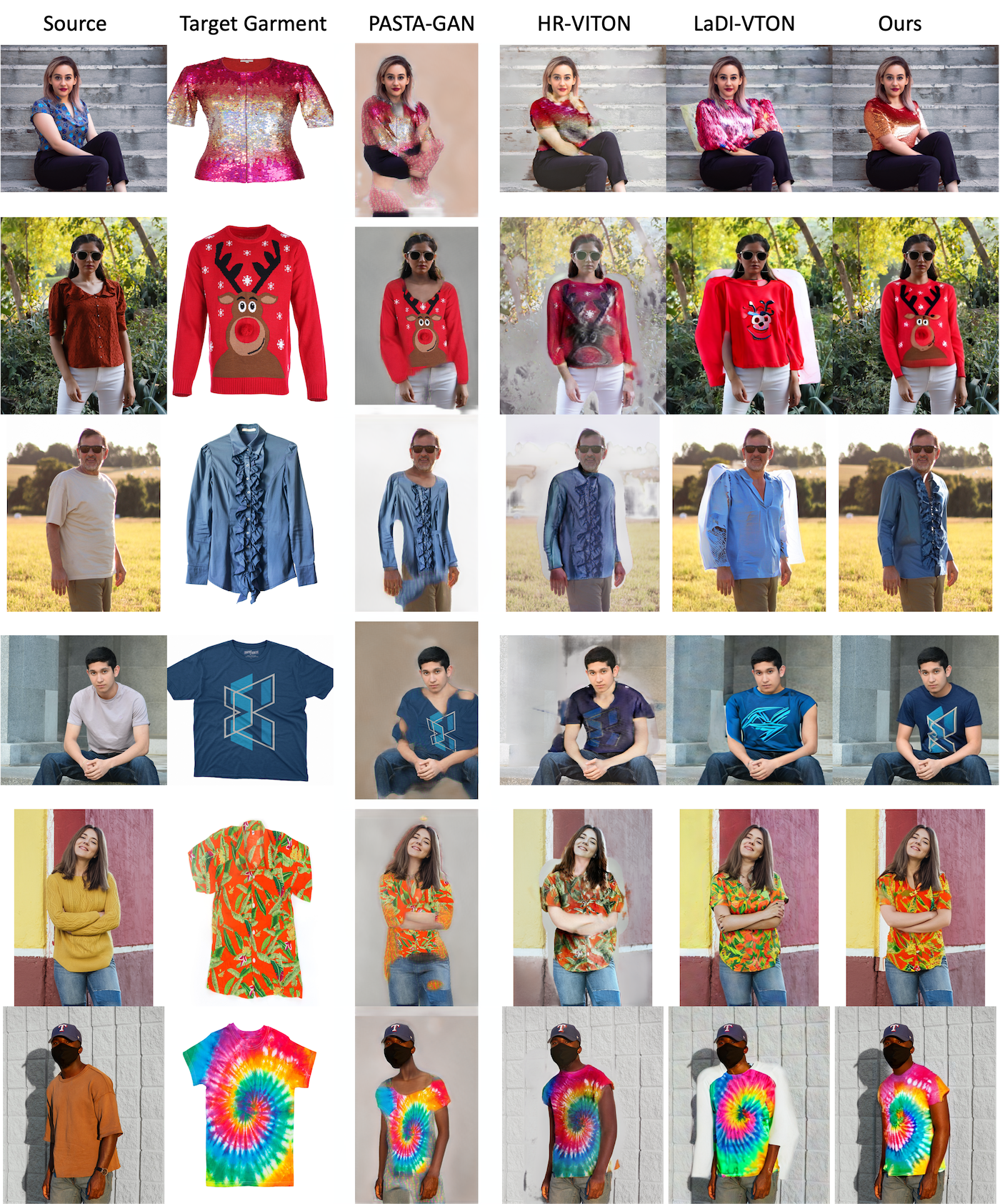}
    \caption{Examples of virtual try-on on WVTON. Our model better fits the new garment onto the person.}
    \label{fig:more_ood_tryon}
\end{figure*}

\cref{fig:more_id_repose} and \cref{fig:more_id_tryon} show in-domain generated results on DeepFashion, DressCode and VITON-HD. \cref{fig:more_ood_repose} and \cref{fig:more_ood_tryon} give several representative examples from our collected WPose and WVTON, respectively. Results show that images generated by our model are better aligned with the target pose while preserving the face and clothing identities.


\clearpage

\clearpage

\clearpage

\clearpage
\begin{figure*}
    \centering
    \includegraphics[width=0.9\textwidth]{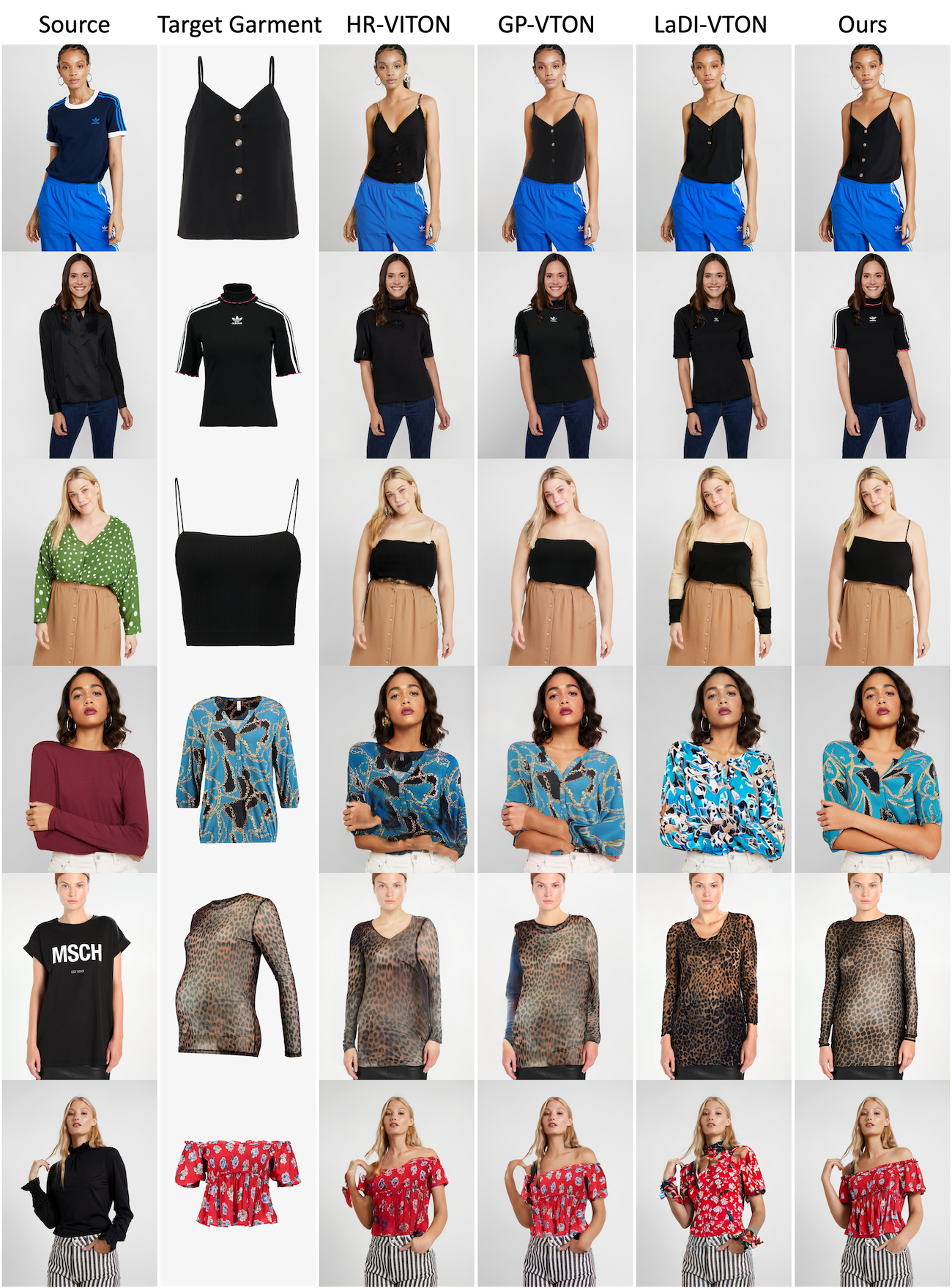}
    \caption{Examples of virtual try-on on VITON-HD. Our model is better at handling occlusions between body parts.}
    \label{fig:more_id_tryon2}
\end{figure*}

\clearpage

\clearpage

\clearpage


\end{document}